\documentclass[10pt]{book}

\usepackage[table]{xcolor}
\usepackage{mathrsfs} 
\usepackage{amssymb,amsmath}
\usepackage{graphicx}
\usepackage{hyperref} 
\usepackage{xspace}
\usepackage{booktabs}
\usepackage{caption} 
\usepackage{longtable}

\usepackage{pdfpages}
\usepackage{caption}

\hypersetup{colorlinks,linkcolor={red!50!black},citecolor={blue!50!black},urlcolor={blue!80!black}}
\usepackage[top=1.2in, bottom=1.2in, left=1.4in, right=1.4in]{geometry}

\usepackage{tikz}
\usetikzlibrary{shapes,calc,positioning,automata,arrows,trees}

\newcommand\bcpen{\includegraphics[width=15pt]{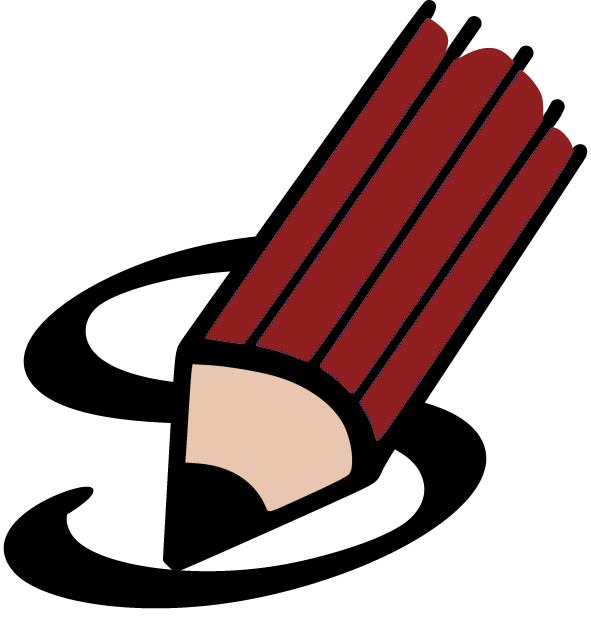}}

\usepackage{listingsutf8}
\usepackage{textcomp}
\usepackage{verbatim}
\usepackage{multirow}

\definecolor{v3lgray}{gray}{0.98}
\definecolor{v2lgray}{gray}{0.85}
\definecolor{vlgray}{gray}{0.92}
\definecolor{dgray}{rgb}{0.4,0.4,0.4}
\definecolor{dblue}{RGB}{0,0,99}
\definecolor{dred}{RGB}{175,6,54}
\definecolor{lred}{RGB}{155,6,44}
\definecolor{dgreen}{RGB}{47,135,7}
\definecolor{dviolet}{RGB}{102,0,153}
\definecolor{mblue}{RGB}{0,0,180}
\definecolor{dorange}{RGB}{204, 82, 0}

\def\ace{ACE\xspace}

\def\x3{{\rm XCSP$^3$}\xspace}
\def\jv3{{\rm JvCSP$^3$}\xspace}
\def\p3{{\rm PyCSP$^3$}\xspace}

\newcommand{\gb}[1]{{\tt #1}} 

\newtheorem{remark}{Remark}

\lstdefinelanguage{json}{
    basewidth  = {.6em,0.6em},
    basicstyle=\normalfont\ttfamily,
    breaklines=true,
    morestring=[b]',
    morestring=[b]", 
    sensitive=false,
    stringstyle=\color[rgb]{0.227,0.226,0.441}\ttfamily, 
    escapechar=!,
    showstringspaces=false,
    xleftmargin=20pt, 
    breaklines=true,basicstyle=\ttfamily\small,inputencoding=utf8/latin9,texcl
}
\lstnewenvironment{json}{\lstset{language=json}}{}

\lstdefinelanguage{mcsp}{
  keywords={forall,array,block,class,implements,model,public,slide},
  basewidth  = {.6em,0.6em},
  keywordstyle=\color{dred}\bfseries,
  ndkeywords={intension,lessThan,lessEqual,greaterEqual,greaterThan,equal,different,implication,equivalence,conjunction,disjunction,extension,regular,mdd,allDifferent,allDifferentMatrix,allEqual,ordered,increasing,decreasing,strictlyIncreasing,strictlyDecreasing,lex,lexMatrix,sum,count,atMost,atLeast,exactly,atMost1,atleast1,exactly1,element,channel,maximum,minimum,cardinality,nValues,noOverlap,cumulative,instantiation,clause,circuit,minimize,maximize},
  ndkeywordstyle=\color{mblue}\bfseries,
  identifierstyle=\color{black},
  sensitive=false,
  comment=[l]{//},
  morecomment=[s]{<!--}{-->},
  commentstyle=\color{dgreen}\ttfamily,
  stringstyle=\color{dgreen}\rmfamily, 
  morestring=[b]',
  morestring=[b]",
  escapechar=~,
  showstringspaces=false,
  classoffset=2, morekeywords={private},keywordstyle=\color{gray},
  classoffset=3, morekeywords={dom,size,when},keywordstyle=\color{dorange},
  xleftmargin=-22pt,xrightmargin=-22pt,
  breaklines=true,basicstyle=\ttfamily\footnotesize,backgroundcolor=\color{v3lgray},inputencoding=utf8/latin9,texcl
}
\lstnewenvironment{mcsp}{\lstset{language=mcsp}}{}

\definecolor{colorex}{RGB}{255,248,220} 
\definecolor{mgray}{rgb}{0.55,0.55,0.55}
\definecolor{officegreen}{rgb}{0.0, 0.5, 0.0}

\newcounter{cntPy}

\lstnewenvironment{python}{\lstset{
    inputencoding=utf8/latin1,
    language=python,
    ndkeywords={size,dom,matrix,strict,occurrences},
    ndkeywordstyle=\color{mblue}, 
    keywords=[2]{data,satisfy,minimize,maximize,solve}, 
    keywordstyle=[2]\color{dred}, 
    deletekeywords={def},
    keywords=[3]{def,If,Then,Else,Match}, 
    keywordstyle=[3]\color{dorange},
    commentstyle=\color{officegreen}, 
    showstringspaces=false,
    captionpos=t,
    basicstyle=\ttfamily\footnotesize,
    framesep=5pt,
    xleftmargin=5pt,xrightmargin=0pt,
    backgroundcolor=\color{colorex}, 
    escapechar=@
  }
}{}

\usepackage[skins,breakable,xparse]{tcolorbox}
\newcommand{\core}[1]{ 
  \medskip \begin{tcolorbox}[
    enhanced,breakable,
    boxsep=0pt,top=4pt,bottom=0pt,left=2mm,right=1mm,
    toprule=0.1mm,leftrule=0.1mm,rightrule=0.25mm,bottomrule=0.25mm,shadow={0.2mm}{-0.2mm}{0mm}{dgray},
    overlay unbroken and first={\node (logo) at ([xshift=4mm,yshift=-5mm]frame.north west) {#1}; },
    colframe=dgray,titlerule=-0.2mm,toptitle=3mm,coltitle=dred, fonttitle=\bfseries,
    lines before break=6, pad at break*=10pt
    }

\newenvironment{boxpy}
 {\stepcounter{cntPy} \core{\bcpen} , colback=colorex, title style={color=colorex}, title=~ ~ \,PyCSP$^3$ Model \thecntPy]}
 {\end{tcolorbox}}

\newenvironment{command}
  {\quote\small\verbatim}
  {\endverbatim\endquote}

  \usepackage{authblk}
 
\title{\textcolor{dred}{Proceedings of the \x3 Competition 2024}}
\author{Gilles Audemard \and Christophe Lecoutre \and Emmanuel Lonca}

\affil{\vspace{1cm} CRIL \\University of Artois \& CNRS \\ France}

\date{November 26, 2024}

\begin{document}
\maketitle

~ \\
~ \\

\bigskip

This document represents the proceedings of the \x3 Competition 2024, following those published in 2022 \cite{compet22} and 2023 \cite{compet23}.
The website containing all {\bf detailed results} of this international competition is available at:
\begin{quote}
  \href{https://www.cril.univ-artois.fr/XCSP24/}{https://www.cril.univ-artois.fr/XCSP24}
\end{quote}

  \bigskip
  \bigskip
\noindent The organization of this 2024 competition involved the following tasks:
\begin{itemize}
\item adjusting general details (dates, tracks, $\dots$) by G. Audemard, C. Lecoutre and E. Lonca
\item selecting instances (problems, models and data) by C. Lecoutre
\item receiving, testing and executing solvers on CRIL cluster by E. Lonca
\item validating solvers and rankings by C. Lecoutre and  E. Lonca
\item developping the 2024 website dedicated to results by G. Audemard
\end{itemize}

\bigskip\bigskip 
{\bf Important}: for reproducing the experiments and results, it is important to use the very same set of \x3 instances, as used in the competition.
These instances can be found in this \href{https://www.cril.univ-artois.fr/~lecoutre/compets/instancesXCSP24.zip}{archive}.
Some (usually minor) differences may exist when compiling the models presented in this document and those that can be found in this  \href{https://www.cril.univ-artois.fr/~lecoutre/compets/modelsXCSP24.zip}{archive}.


\tableofcontents

\chapter{About the Selection of Problems in 2024}

Remember that the complete description, {\bf Version 3.2}, of the format (\x3) used to represent combinatorial constrained problems can be found in \cite{xcsp3}.
As usual for \x3 competitions, we have limited \x3 to its kernel, called \x3-core \cite{xcsp3core}.
This means that the scope of the \x3 competition is restricted to:
\begin{itemize}
\item integer variables,
\item CSP (Constraint Satisfaction Problem) and COP (Constraint Optimization Problem), 
\item a set of 24 popular (global) constraints for main tracks:
  \begin{itemize}
  \item generic constraints: \gb{intension} and \gb{extension} (also called \gb{table})
  \item language-based constraints: \gb{regular} and \gb{mdd}
  \item comparison constraints: \gb{allDifferent}, \gb{allDifferentList}, \gb{allEqual}, \gb{ordered}, \gb{lex} and \gb{precedence}
  \item counting/summing constraints: \gb{sum}, \gb{count}, \gb{nValues} and \gb{cardinality}
  \item connection constraints: \gb{maximum}, \gb{minimum}, \gb{element} and \gb{channel}
  \item packing/scheduling constraints: \gb{noOverlap}, \gb{cumulative}, \gb{binPacking} and \gb{knapsack}
  \item \gb{circuit}, \gb{instantiation} and \gb{slide}
  \end{itemize}
  and a smaller set of constraints for mini tracks.
\end{itemize}

\medskip
For the 2024 competition, 34 problems have been selected.
They are succinctly presented in Table \ref{fig:problems}.
For each problem, the type of optimization is indicated (if any), as well as the involved constraints.
At this point, do note that making a good selection of problems/instances is a difficult task.
In our opinion, important criteria for a good selection are:
\begin{itemize}
\item the novelty of problems, avoiding constraint solvers to overfit already published problems;
\item the diversity of constraints, trying to represent all of the most popular constraints (those from \x3-core) while paying attention to not over-representing some of them;
\item the scaling up of problems.
\end{itemize}

\begin{table}
  \begin{tabular}{p{5.5cm}p{0.2cm}p{7.5cm}}
    \toprule
CSP Problems & & Global Constraints\\
    \midrule
\rowcolor{vlgray}{}  AverageAvoiding & & \gb{allDifferent}, \gb{minimum}  \\
    FastMatrixMultiplication &   & \gb{lex},  \gb{precedence}, \gb{sum}, \gb{table$^*$} \\
\rowcolor{vlgray}{}  Fillomino   &  & \gb{element}, \gb{sum}, \gb{table$^*$}  \\
    FRB & &  \gb{table}  \\
\rowcolor{vlgray}{}   Hamming & &  \gb{lex}, \gb{sum} \\
    HyperSudoku & & \gb{allDifferent} \\
\rowcolor{vlgray}{}  MisteryShopper & & \gb{allDifferent}, \gb{channel}, \gb{lex}, \gb{table}\\
Pentominoes & &  \gb{allDifferent}, \gb{table} \\
\rowcolor{vlgray}{} PoolBallTriangle & & \gb{allDifferent} \\
RotatingWorkforce2 & & \gb{cardinality}, \gb{count}, \gb{regular}, \gb{sum} \\
\rowcolor{vlgray}{} Soccer & & \gb{allDifferent}, \gb{sum}, \gb{table} \\SocialGolfers &  & \gb{allDifferent}, \gb{cardinality}, \gb{lex}, \gb{sum} \\
\rowcolor{vlgray}{} SolitairePattern & & \gb{table} \\
Subisomorphism &  & \gb{allDifferent}, \gb{table} \\
\rowcolor{vlgray}{} Takuzu & & \gb{AllDifferentList}, \gb{sum} \\
WordSquare &  & \gb{allDifferentList}, \gb{element}, \gb{table} \\
& & \\
\midrule
COP Problems & & Global Constraints \\ 
\midrule
\rowcolor{vlgray}{} AircraftAssemblyLine &   & \gb{cumulative}, \gb{noOverlap}, \gb{sum} \\
 AztecDiamondSym &   &  \gb{cardinality}, \gb{sum}, \gb{table$^*$} \\
\rowcolor{vlgray}{} BinPacking &   & \gb{binPacking}, \gb{cardinality}, \gb{nValues}, \gb{lex}, \gb{sum} \\ 
Cargo &  & \gb{cumulative}, \gb{element}, \gb{noOverlap}, \gb{sum} \\
\rowcolor{vlgray}{} Charlotte &   & \gb{allDifferent}, \gb{count}, \gb{sum}, \gb{table$^*$} \\
Drinking &  & \gb{sum}, \gb{table} \\
\rowcolor{vlgray}{} FoolSolitaire &  & \gb{element}, \gb{table} \\
LitPuzzle &  & \gb{sum}  \\
\rowcolor{vlgray}{} MaxDensityOscillatingLife &  & \gb{lex}, \gb{sum}, \gb{table$^*$} \\
Pyramid &   & \gb{allDifferent}  \\
\rowcolor{vlgray}{} RubiksCube &  & \gb{allDifferent}, \gb{element} \\
SameQueensKnights &  & \gb{sum} \\
\rowcolor{vlgray}{} StillLife &  & \gb{sum}, \gb{table} \\
TestScheduling &  &  \gb{cumulative}, \gb{noOverlap}, \gb{maximum} \\
\rowcolor{vlgray}{} TravelingTournament &   & \gb{allDifferent}, \gb{cardinality}, \gb{element}, \gb{regular}, \gb{sum}, \gb{table$^*$} \\
VRP\_LC &  & \gb{circuit}, \gb{cumulative}, \gb{element}, \gb{sum}  \\
\rowcolor{vlgray}{} WordGolf & & \gb{element}, \gb{sum} \\
Wordpress &  & \gb{element}, \gb{sum} \\
\bottomrule
  \end{tabular}
   \caption{Selected Problems for the main tracks of the 2024 Competition. When \gb{table} is followed by ($*$), it means that starred tables are involved.}\label{fig:problems}

\end{table}

\paragraph{Novelty.} Almost all problems are new in 2024, with models directly written in \p3. One problem series has been submitted in response to the call (AircraftAssemblyLine, as in 2023). 


\paragraph{Scaling up.} It is always interesting to see how constraint solvers behave when the instances of a problem become harder and harder.
This is what we call the scaling behavior of solvers.
For most of the problems in the 2024 competition, we have selected series of instances with regular increasing difficulty.
It is important to note that assessing the difficulty of instances was mainly determined with \ace \cite{ace}, which is the reason why \ace is declared to be off-competition (due to this strong bias).

\paragraph{Selection.} This year, the selection of problems and instances has been performed by Christophe Lecoutre.
As a consequence, the solver \ace was labeled off-competition.


\bigskip

\chapter{Problems and Models}

In the next sections, you will find all models used for generating the \x3 instances of the 2024 competition (for main CSP and COP tracks).
Almost all models are written in \p3 \cite{pycsp3}, Version 2.4 (officially released in August 28, 2024); see \href{https://pycsp.org}{https://pycsp.org}.

\section{CSP}

\subsection{Average Avoiding}

\paragraph{Description.}
Arrange an array such that the average of 2 numbers does not lie between them in the array.
It seems that it was a Facebook interview question, and a question on StackOverflow.

\paragraph{Data.}

Because, here, we consider to find an array with all values ranging (in some order) from 0 to a specified limit, only one integer is required to specify a specific instance: the order $n$ of the array.
The values of $n$ used for generating the 2024 competition instances are:
\begin{quote}
20, 25, 30, 35, 40, 45, 50, 55, 60, 65
\end{quote}

\paragraph{Model.}
The \p3 model, in a file `AverageAvoiding.py', used for the competition is: 

\begin{boxpy}\begin{python}
@\imp@

n = data 

# x[i] is the ith value of the sequence
x = VarArray(size=n, dom=range(n))

satisfy(
   # ensuring that the average of 2 numbers of the sequence separated by a value v is not v
   [2 * x[k] != x[i] + x[j]
      for i, j in combinations(n, 2) if i + 1 < j for k in range(i + 1, j)],

   # ensuring that we have a permutation of the initial sequence
   AllDifferent(x),
  
   # tag(symmetry-breaking)
   x[0] == Minimum(x)
)
\end{python}\end{boxpy}

This model involves a one-dimensional array of variables $x$, a constraint \gb{AllDifferent}, a constraint \gb{Minimum} and a group of intensional constraints.
A series of 10 instances has been selected for the competition.
For generating an \x3 instance (file), you can execute for example:
\begin{command}
python AverageAVoiding.py -data=20
\end{command}

\begin{remark}
In such a restricted configuration (all values ranging from 0 to $n-1$), one could have written \verb! x[0] == 0! instead of posting a constraint \gb{Minimum}.
Besides, it is not very clear that it is a correct way of breaking symmetries (in the general case, this should be demonstrated).   
\end{remark}

\subsection{Fast Matrix Multiplication}

\paragraph{Description.}

Taken from \cite{DLVK_fast}: 
``The multiplication of two matrices $A$ and $B$ of sizes $n \times m$ and $m \times p$ results in a product matrix $C$ of size $n \times p$.
This operation can be represented by a binary third-order tensor $T$.
An entry $T_{i,j,k}$ of this tensor is equal to 1 if and only if the kth entry in the output matrix $C$ uses the scalar product of the ith entry of $A$ and the jth entry of $B$.
The FMM (Fast Matrix Multiplication) problem for a given tensor $T$, rank $R$, and field $F$ (e.g., $F = \{-1, 0, +1\}$) asks: can each entry $T_{i,j,k}$ of $T$ be expressed as the sum of exactly $R$ trilinear terms involving the factor matrices $U$, $V$, and $W$, as follows:

$T_{i,j,k} =  \Sigma_{r=1}^{R} U_{i,r} \times V_{j,r} \times W_{k,r}, \; \forall i \in \{1, \dots, n \times m \}, j \in \{1, \dots, m \times p \}, k \in \{1, \dots, n \times p \}$''

\paragraph{Data.}

Four integers are required to specify a specific instance. Values of $(n,m,p,R)$ used for the instances in the competition are:
\begin{quote}
(2,2,2,3), (2,2,2,4), (2,2,2,5), (2,2,2,6), (2,2,2,7), (1,3,3,9), (3,1,3,9), (2,2,3,11), (2,3,2,11)
\end{quote}

\paragraph{Model.}
The \p3 model, in `FastMatrixMultiplication.py', used for the competition is: 

\begin{boxpy}\begin{python}
@\imp@

n, m, p, R = data 
U, V, W = n * m, m * p, n * p

T = []
for k in range(n * p):
   M = [[0] * V for _ in range(U)]
   i, j = k // p, k % p
   for a in range(m):
      M[i * m + a][j + (p * a)] = 1
   T.append(M)

# x is the first factor matrix
x = VarArray(size=[U, R], dom={-1, 0, 1})

# y is the second factor matrix
y = VarArray(size=[V, R], dom={-1, 0, 1})

# z is the third factor matrix
z = VarArray(size=[W, R], dom={-1, 0, 1})

satisfy(
   # ensuring that the tensor is produced
   [Sum(x[i][r] * y[j][r] * z[k][r] for r in range(R)) == T[k][i][j]
      for i in range(U) for j in range(V) for k in range(W)],

   # tag(symmetry-breaking)
   [
      [LexIncreasing(x[:, r] + y[:, r], x[:, r + 1] + y[:, r + 1]) for r in range(R - 1)],
    
      [Precedence(x[:, r], values=[-1, 1]) for r in range(R)],
    
      [Precedence(z[:, r], values=[-1, 1]) for r in range(R)]
   ]
)
\end{python}\end{boxpy}

This model involves three two-dimensional arrays of variables $x$, $y$ and $z$, an (global) constraints of type \gb{Lex}, \gb{Precedence} and \gb{Sum}.
A series of 9 instances has been selected for the competition.
For generating an \x3 instance (file), you can execute for example:
\begin{command}
python FastMatrixMultiplication.py -data=[2,2,2,3]
\end{command}

\subsection{Fillomino}

\paragraph{Description.}
From \href{https://en.wikipedia.org/wiki/Fillomino}{Wikipedia}: ``Fillomino is played on a rectangular grid.
Some cells of the grid start containing numbers.
The goal is to divide the grid into regions called polyominoes (by filling in their boundaries) such that each given number n in the grid satisfies the following constraints:
i) each clue n is part of a polyomino of size n; ii) no two polyominoes of matching size (number of cells) are orthogonally adjacent (share a side).''

\paragraph{Data.}

As an illustration of data specifying an instance of this problem, we have:
\begin{json}
{
  "puzzle": [
    [4, 0, 0, 3, 0],
    [0, 5, 4, 0, 2],
    [0, 6, 1, 0, 1],
    [0, 0, 0, 6, 0],
    [5, 0, 0, 3, 0]
  ]
}
\end{json}

\paragraph{Model.}
The \p3 model, in a file `Fillomino.py', used for the competition is: 

\begin{boxpy}\begin{python}
@\imp@

puzzle = data
n, m = len(puzzle), len(puzzle[0])

Values = {puzzle[i][j] for i in range(n) for j in range(m) if puzzle[i][j] > 0}

seen = set()
preassigned = []
for i in range(n):
   for j in range(m):
      if puzzle[i][j] == 1 or (puzzle[i][j] > 1 and puzzle[i][j] not in seen): 
         preassigned.append((len(preassigned), puzzle[i][j], i + 1, j + 1))  # +1 for border
         seen.add(puzzle[i][j])

remaining = n * m - sum(sz for (_, sz, _, _) in preassigned)
nRegions = len(preassigned) + remaining  
maxDistance = max(max(Values), remaining) + 1  

Points = [(i, j) for i in range(1, n + 1) for j in range(1, m + 1)]

def tables():
   Tc = {(1, ANY, ANY, ANY, ANY, ANY, 0, ANY, ANY, ANY, ANY)}
   for k in range(nRegions):  # root forced to be the first cell of the region wrt top left
      Tc.add((gt(1), k, ANY, k, ANY, ANY, 0, ANY, 1, ANY, ANY))
      Tc.add((gt(1), k, ANY, ANY, ANY, k, 0, ANY, ANY, ANY, 1))
      for v in range(1, maxDistance):
         Tc.add((gt(1), k, k, ANY, ANY, ANY, v, v - 1, ANY, ANY, ANY))
         Tc.add((gt(1), k, ANY, k, ANY, ANY, v, ANY, v - 1, ANY, ANY))
         Tc.add((gt(1), k, ANY, ANY, k, ANY, v, ANY, ANY, v - 1, ANY))
         Tc.add((gt(1), k, ANY, ANY, ANY, k, v, ANY, ANY, ANY, v - 1))
   Tr = {(v, v, k, k) for v in Values for k in range(nRegions)}
         | {(v, ne(v), k, ne(k)) for v in Values for k in range(nRegions)}   
   return Tc, Tr

Tc, Tr = tables()  # tables for connections and regions

# x[i][j] is the region (number) where the square at row i and column j belongs
x = VarArray(size=[n + 2, m + 2], dom=range(nRegions), dom_border={-1})

# y[i][j] is the value of the square at row i and column j
y = VarArray(size=[n + 2, m + 2], dom=lambda i,j: {puzzle[i-1][j-1]} if puzzle[i-1][j-1] != 0 else Values, dom_border={-1})

# d[i][j] is the distance of the square at (i,j) wrt the starting square of the (same) region
d = VarArray(size=[n + 2, m + 2], dom=range(maxDistance), dom_border={-1})

# s[k] is the size of the kth region
s = VarArray(size=nRegions, dom={0} | Values)  # it is important to have 0

satisfy(
   # setting starting squares of pre-assigned regions
   [
      (
         x[i][j] == k,
         y[i][j] == sz,
         s[k] == sz
      ) for k, sz, i, j in preassigned
   ],

   # setting values according to the size of the regions
   [y[i][j] == s[x[i][j]] for i, j in Points if puzzle[i - 1][j - 1] == 0],

   # controlling the size of each region
   [s[k] == Sum(x[i][j] == k for i, j in Points) for k in range(nRegions)],
   
   # ensuring connection
   [(y[i][j], x.cross(i, j), d.cross(i, j)) in Tc for i, j in Points],
   
   # two regions of the same size cannot have neighbouring squares
   [
      [(y[i][j], y[i][j + 1], x[i][j], x[i][j + 1]) in Tr
         for i in range(1, n + 1) for j in range(1, m)],

      [(y[i][j], y[i + 1][j], x[i][j], x[i + 1][j]) in Tr
         for j in range(1, m + 1) for i in range(1, n)]
   ],

   # ensuring a single root for each region  tag(symmetry-breaking)
   [
      If(
         d[i][j] == 0, d[p][q] == 0,
         Then=x[i][j] != x[p][q]
      ) for (i, j) in Points for (p, q) in Points if (i, j) != (p, q)
   ],
  
   # pushing regions of size 0 on the right  tag(symmetry-breaking)
   [
      If(
         s[k] == 0,
         Then=s[k + 1] == 0
      ) for k in range(nRegions - 1)
   ]
)
\end{python}\end{boxpy}

This model involves four arrays of variables, and (global) constraints of type \gb{Element}, \gb{Sum} and \gb{Table}.

A series of 10 instances has been selected for the competition.
For generating an \x3 instance (file), you can execute for example:
\begin{command}
python Fillomino.py -data=grid.json
\end{command}
where `grid.json' is a data file in JSON format.

\subsection{FRB}

This problem has already been selected in previous XCSP competitions.
This satisfaction problem only involves (ordinary) table constraints.
A series of 12 instances has been selected for the competition.
These instances were randomly generated using Model RB \cite{XBHL_random}, while guaranteeing satisfiability.

\subsection{Hamming}

\paragraph{Description.}

Given four integers $n$, $m$, $d$ and $k$, the goal is to find $n$ vectors of size $m$ where each value lies between $0$ and $d$ (exclusive), and every two vectors have a Hamming distance at most equal to $k$.
See \href{https://en.wikipedia.org/wiki/Hamming_distance}{Wikipedia}.

\paragraph{Data.}

Four integers are required to specify a specific instance. Values of $(n,m,d,k)$ used for the instances in the competition are:
\begin{quote}
(20,10,3,5), (20,10,3,6), (20,10,3,7), (20,10,3,8), (20,10,5,7), (20,10,5,8), (20,10,5,9), (20,10,5,10), (30,15,7,12), (30,15,7,13), (30,15,7,14), (30,15,7,15)
\end{quote}

\paragraph{Model.}
The \p3 model, in a file `Hamming.py', used for the competition is: 

\begin{boxpy}\begin{python}
@\imp@

n, m, d, k = data or (9, 4, 3, 3)

# x[i][j] is the jth value of the ith vector
x = VarArray(size=[n, m], dom=range(d))

satisfy(
   # ensuring a Hamming distance of at least 'k' between any two vectors
   [Hamming(row1, row2) >= k for row1, row2 in combinations(x, 2)],

   # tag(symmetry-breaking)
   LexIncreasing(x)
)
\end{python}\end{boxpy}

This model involves a two-dimensional array of variables $x$, a group of constraints \gb{Sum} (derived from the \gb{Hamming} function), and a constraint \gb{Lex}.

A series of 12 instances has been selected for the competition.
For generating an \x3 instance (file), you can execute for example:
\begin{command}
python Hamming.py -data=[20,10,3,5]
\end{command}

\subsection{Hyper Sudoku}

\paragraph{Description.}

Hyper Sudoku differs from Sudoku by having additional constraints.
When the base of the grid is 3 (as usually for Sudoku) there are four 3-by-3 blocks in addition to the major 3-by-3 blocks that also require exactly one entry of each numeral from 1 through 9.

\paragraph{Data.}

In our model, only one integer is required to specify a specific instance: the base of the grid.
The values used for generating the 2024 competition instances are:
\begin{quote}
3, 4, 5, 6, 7, 8, 9, 10
\end{quote}

\paragraph{Model.}
The \p3 model, in a file `HyperSudoku.py', used for the competition is: 

\begin{boxpy}\begin{python}
@\imp@

base = data  # (base of the grid)
n = base * base

# x[i][j] is the value of cell with coordinates (i,j)
x = VarArray(size=[n, n], dom=range(1, n + 1))

satisfy(
   # imposing distinct values on each row and each column
   AllDifferent(x, matrix=True),

   # imposing distinct values on (main) blocks
   [AllDifferent(x[i:i + base, j:j + base])
      for i in range(0, n, base) for j in range(0, n, base)],
  
   # imposing distinct values on shaded blocks
   [AllDifferent(x[i:i + base, j:j + base])
      for i in range(1, n - base, base + 1) for j in range(1, n - base, base + 1)]
)
\end{python}\end{boxpy}

This model involves a two-dimensional array of variables $x$, and many global constraints \gb{AllDifferent} (one of them in matrix form).
A series of 8 instances has been selected for the competition.
For generating an \x3 instance (file), you can execute for example:
\begin{command}
python HyperSudoku.py -data=8
\end{command}

\subsection{Mistery Shopper}

This is Problem \href{https://www.csplib.org/Problems/prob004}{004} on CSPLib.

\paragraph{Description.}

From Jim Ho Man Lee: ``A well-known cosmetic company wants to evaluate the performance of their sales people, who are stationed at the company’s counters at various department stores in different geographical locations. For this purpose, the company has hired some secret agents to disguise themselves as shoppers to visit the sales people. The visits must be scheduled in such a way that each sales person must be visited by shoppers of different varieties and that the visits should be spaced out roughly evenly. Also, shoppers should visit sales people in different geographic locations.''
The complete description can be found at CSPLib.

\paragraph{Data.}

Five integers are required to specify a specific random instance, generated with the help of the file `MisteryShopper\_Random.py'. Values of $(g,n,kr,ke,s)$ used for generating the instances in the competition are:
\begin{quote}
(4,6,1,6,0), (4,6,2,6,0), (5,8,1,6,0), (5,8,2,6,0), (5,10,0,6,0), (5,10,1,6,0), (5,10,2,6,0), (5,10,3,6,0), (6,12,0,6,0), (6,12,1,6,0), (6,12,2,6,0), (6,12,3,6,0)
\end{quote}

\paragraph{Model.}
The \p3 model, in a file `MisteryShopper.py', used for the competition is: 

\begin{boxpy}\begin{python}
@\imp@

vr_sizes = data.visitorGroups  # vr_sizes[i] gives the size of the ith visitor group
ve_sizes = data.visiteeGroups  # ve_sizes[i] gives the size of the ith visitee group

nVisitors, nVisitees = sum(vr_sizes), sum(ve_sizes)
if nVisitors - nVisitees > 0:
   ve_sizes.append(nVisitors - nVisitees)  # an artificial group with dummy visitees is added
n, nWeeks = nVisitors, len(vr_sizes)

vr_table = {(i, sum(vr_sizes[:i])+j) for i, size in enumerate(vr_sizes) for j in range(size)}
ve_table = {(i, sum(ve_sizes[:i])+j) for i, size in enumerate(ve_sizes) for j in range(size)}

# r[i][w] is the visitor for the ith visitee at week w
r = VarArray(size=[n, nWeeks], dom=range(n))

# e[i][w] is the visitee for the ith visitor at week w
e = VarArray(size=[n, nWeeks], dom=range(n))

# rg[i][w] is the visitor group for the ith visitee at week w
rg = VarArray(size=[n, nWeeks], dom=range(len(vr_sizes)))

# eg[i][w] is the visitee group for the ith visitor at week w
eg = VarArray(size=[n, nWeeks], dom=range(len(ve_sizes)))

satisfy(
   # each week, all visitors must be different
   [AllDifferent(col) for col in columns(r)],

   # each week, all visitees must be different
   [AllDifferent(col) for col in columns(e)],
  
   # the visitor groups must be different for each visitee
   [AllDifferent(row) for row in rg],
   
   # the visitee groups must be different for each visitor
   [AllDifferent(row) for row in eg],
   
   # channeling arrays vr and ve, each week
   [Channel(r[:, w], e[:, w]) for w in range(nWeeks)],
   
   # tag(symmetry-breaking)
   [
      LexIncreasing(r, matrix=True),
    
      [Increasing(r[nVisitees:n, w], strict=True) for w in range(nWeeks)]
   ],
  
   # linking a visitor with its group
   [(rg[i][w], r[i][w]) in vr_table for i in range(n) for w in range(nWeeks)],
   
   # linking a visitee with its group
   [(eg[i][w], e[i][w]) in ve_table for i in range(n) for w in range(nWeeks)]
)
\end{python}\end{boxpy}

This model involves four two-dimensional arrays of variables, and (global) constraints of type \gb{AllDifferent}, \gb{Channel}, \gb{Lex} and \gb{Table}.
A series of 12 instances has been selected for the competition.
For generating an \x3 instance (file), you can execute for example:
\begin{command}
python MisteryShopper.py -parser=MisteryShopper_Random.py 4 6 1 6 0 
\end{command}

\subsection{Pentominoes}

\paragraph{Description.}

Here, the goal is to fill a grid of size $n$ by $m$ with different pentominoes.
See \href{https://en.wikipedia.org/wiki/Pentomino}{Wikipedia}.

\paragraph{Data.}

Two integers are required to specify a specific instance. Values of $(n,m)$ used for generating the instances in the competition are:
\begin{quote}
(3,20), (4,15), (5,12), (6,10), (3,30), (5,18), (6,15), (9,10), (3,40), (4,30), (5,24), (6,20), (8,15), (10,12)
\end{quote}

\paragraph{Model.}
The \p3 model, in a file `Pentominoes.py', used for the competition is: 

\begin{boxpy}\begin{python}
@\imp@

n, m = data
assert n * m in (60, 90, 120)  # for the moment
nPieces = (n * m) // 5

pentominoes = list(polyominoes[5].values())  # 12 distinct pentominoes

def table(pentomino):
   T = []
   for i in range(n):
      for j in range(m):
         for sym in TypeSquareSymmetry.symmetric_patterns(pentomino):
            if all(0 <= i + k < n and 0 <= j + l < m for k, l in sym):
               T.append(tuple((i + k) * m + (j + l) for k, l in sym))
   return T


# x[i][j] is the board cell index where is put the jth piece of the ith pentomino
x = VarArray(size=[nPieces, 5], dom=range(n * m))

satisfy(
   # positioning all pentominoes correctly
   [x[i] in table(pentominoes[i % 12]) for i in range(nPieces)],

   # ensuring no overlapping pieces
   AllDifferent(x),
       
   # tag(symmetry-breaking)
   [x[i % 12][0] < x[i][0] for i in range(13, nPieces)]
)
\end{python}\end{boxpy}

This model involves a two-dimensional array of variables, and (global) constraints of type \gb{AllDifferent} and \gb{Table}.
A series of 14 instances has been selected for the competition.
For generating an \x3 instance (file), you can execute for example:
\begin{command}
python Pentominoes.py -data=[6,20]
\end{command}

\subsection{PoolBall Triangle}

\paragraph{Description.}

From Martin Gardner: ``Given n*(n+1) / 2 numbered pool balls in a triangle, is it  possible to place them so that the number of each ball below  two balls is the difference of (the number of) those two balls?''

\paragraph{Data.}

In our model, only one integer is required to specify a specific instance: the value of $n$.
The values used for generating the 2024 competition instances are:
\begin{quote}
5, 7, 10, 11, 12, 13, 14, 15, 16, 18, 20

\end{quote}

\paragraph{Model.}
The \p3 model, in a file `PoolBallTriangle.py', used for the competition is: 

\begin{boxpy}\begin{python}
@\imp@

n = data 
k = (n * (n + 1)) // 2

x = VarArray(size=[n, n], dom=lambda i, j: range(1, k + 1) if i < n - j else None)

satisfy(
   AllDifferent(x),

   [x[i][j] == abs(x[i - 1][j] - x[i - 1][j + 1]) for i in range(1, n) for j in range(n-i)],

   # tag(symmetry-breaking)
   x[-2][0] < x[-2][1]
)
\end{python}\end{boxpy}

This model involves a two-dimensional array of variables $x$, and the global constraint \gb{AllDifferent}.
A series of 11 instances has been selected for the competition.
For generating an \x3 instance (file), you can execute for example:
\begin{command}
python PoolBallTriangle.py -data=10
\end{command}

\subsection{Rotating Workforce}

\paragraph{Description.}

From \cite{MSS_solver}: ``Rotating workforce scheduling is a specific personnel scheduling problem arising in many spheres of life such as, e.g., industrial plants, hospitals, public institutions, and airline companies.
A schedule must meet many constraints such as workforce requirements for shifts and days, minimal and maximal length of shifts, and shift transition constraints.''

\paragraph{Data.}

Two integers are required to specify a specific random instance, generated with the help of the file `RotatingWorkforce\_Random.py'. Values of $n,s)$ used for generating the instances in the competition are:
\begin{quote}
(25,0), (25,1), (25,2), (40,0), (40,1), (40,2), (60,0), (60,1), (60,2), (80,0), (80,1), (80,2), (100,0), (100,1), (100,2) 
\end{quote}

\paragraph{Model.}
The \p3 model, in a file `RotatingWorkforce2.py', used for the competition is: 

\begin{boxpy}\begin{python}
@\imp@

nEmployees, requirements = data
nWeeks, nDays = nEmployees, 7

SATURDAY, SUNDAY = (5, 6)  # weekend
OFF, DAY, EVENING, NIGHT = Shifts = range(4)
nShifts = len(Shifts)
NOT_OFF = range(1, nShifts)

q0, q1, q2, q3, q4 = states = Automaton.states_for(range(5))

A1 = Automaton(start=q0, final=q2, transitions=[(q0, NOT_OFF, q0), (q0, OFF, q1),
       (q1, NOT_OFF, q0), (q1, OFF, q2), (q2, Shifts, q2)])

A2 = Automaton(start=q0, final=states, transitions=[(q0, OFF, q0), (q0, (DAY, EVENING), q4),
       (q0, NIGHT, q1), (q1, OFF, q0), (q1, NIGHT, q2), (q2, OFF, q0), (q2, NIGHT, q3),
       (q3, OFF, q0), (q4, OFF, q0), (q4, (DAY, EVENING), q4)])


# x[i][j] is the shift for the jth day of the ith week
x = VarArray(size=[nWeeks, nDays], dom=range(nShifts))

# fw[i] is 1 if the weekend of the ith week is free
fw = VarArray(size=nWeeks, dom={0, 1})

x_flat = flatten(x)

satisfy(

   # ensuring shift requirements are met for each day
   [
      Cardinality(
         within=x[:, j],
         occurrences={i: requirements[j][i - 1] for i in range(1, nShifts)}
      ) for j in range(nDays)
   ],
  
   # at least 2 consecutive days off each week
   [x[i] in A1 for i in range(nWeeks)],
  
   # at most five days without rest
   [
      Sum(
         x_flat[j] == 0 for j in range(i, i + 6)
      ) != 0 for i in range(0, nWeeks * nDays)
   ],
  
   # computing free weekends
   [
      fw[i] == both(
         x[i][SATURDAY] == 0,
         x[i][SUNDAY] == 0
      ) for i in range(nWeeks)
   ],
  
   # at least 1 out of 3 weekends off
   [Exist(fw[i:i + 3], value=1) for i in range(nWeeks)],
  
   # at most 3 night-shifts in a row, and then rest
   x[:nWeeks + 2] in A2
)
\end{python}\end{boxpy}

This model involves two arrays of variables, and (global) constraints of type \gb{Cardinality}, \gb{Regular}, and \gb{Sum}.
A series of 15 instances has been selected for the competition.
For generating an \x3 instance (file), you can execute for example:
\begin{command}
python RotatingWorkforce2.py -parser=RotatingWorkforce_Random.py 25 0
\end{command}

\subsection{Soccer}

\paragraph{Description.}

From \cite{DDa_sabio}: ``A soccer competition consists of n teams playing against each other in a single or double round-robin schedule.
The elimination problem is well-known in sports competitions and consists in determining whether at some stage of the competition a given team still has the opportunity to be within the top teams to qualify for playoﬀs or become the champion.
This problem was proved NP-Complete for the current FIFA score system (0 points-loss, 1 point-tie, 3 points-win).
The guaranteed qualiﬁcation problem is to ﬁnd the minimum number of points a given team has to win to become champion or qualify to the playoﬀs.''

Note that SABIO (Soccer Analysis Based on Inference Outputs) is an online platform, available at \href{www.sabiofutbol.com}{www.sabiofutbol.com}, capable of answering soccer related queries by letting users formulate questions in form of constraints.

\paragraph{Data.}

\vspace{-0.2cm}
As an illustration of data specifying an instance of this problem, we have:
\begin{json}
{
  "games": [
    [1, 0],
    [1, 0],
    ...,
    [21, 20]
  ],
  "initial_points": [12, 12, ..., 25],
  "positions": [
    [0, 3],
    [1, 22],
    ...,
    [21, 10]
  ]
}
\end{json}

\paragraph{Model.}

An original model was proposed by Robinson Duque, Alejandro Arbelaez, and Juan Francisco Díaz for the 2018 Minizinc challenge.
The \p3 model, in a file `Soccer.py', used for the \x3 competition is close to (can be seen as the close translation of) it: 

\begin{boxpy}\begin{python}
@\imp@

games, iPoints, positions = data
nGames, nTeams, nPositions = len(games), len(iPoints), len(positions)

P = [0, 1, 3]

lb_score = min(iPoints[i] + sum(min(P) for j in range(nGames) if i in games[j])
                 for i in range(nTeams))
ub_score = max(iPoints[i] + sum(max(P) for j in range(nGames) if i in games[j])
                 for i in range(nTeams))

# points[j] are the points for the two teams (indexes 0 and 1) of the jth game
points = VarArray(size=[nGames, 2], dom=P)

# score[i] is the final score of the ith team
score = VarArray(size=nTeams, dom=range(lb_score, ub_score + 1))

# fp[i] is the final position of the ith team
fp = VarArray(size=nTeams, dom=range(1, nTeams + 1))

# bp[i] is the best possible position of the ith team
bp = VarArray(size=nTeams, dom=range(1, nTeams + 1))

# wp[i] is the worst possible position of the ith team
wp = VarArray(size=nTeams, dom=range(1, nTeams + 1))

satisfy(
   # assigning rights points for each game
   [points[j] in {(0, 3), (1, 1), (3, 0)} for j in range(nGames)],
  
   # computing final points
   [
      score[i] == iPoints[i] + S for i in range(nTeams)
         if (S := Sum(points[j][game.index(i)] for j, game in enumerate(games) if i in game),)
   ],
  
   # computing worst positions (the number of teams with greater total points)
   [
      wp[i] == Sum(score[j] >= score[i] for j in range(nTeams)) for i in range(nTeams)
   ],
  
   # computing best positions from worst positions and the number of teams with equal points
   [
      bp[i] == wp[i] - S for i in range(nTeams)
         if (S := Sum(score[j] == score[i] for j in range(nTeams) if i != j),)
   ],
  
   # bounding final positions
   [
      (
         fp[i] >= bp[i],
         fp[i] <= wp[i]
      ) for i in range(nTeams)
   ],
  
   # ensuring different positions
   AllDifferent(fp),
  
   # applying rules from specified positions
   [
      (
         fp[i] == p,
         Sum(score[j] > score[i] for j in range(nTeams) if j != i) < p,
         Sum(score[j] < score[i] for j in range(nTeams) if j != i) <= nTeams + 1 - p
      ) for i, p in positions
   ]
)
\end{python}\end{boxpy}

This model involves five arrays of variables, and (global) constraints of type \gb{AllDifferent}, \gb{Sum}, and \gb{Table}.
A series of 16 instances has been selected for the competition (files have been gently provided by Robinson Duque). 
For generating an \x3 instance (file), you can execute for example:
\begin{command}
python Soccer.py -data=inst.json
\end{command}
where `inst.json' is a data file in JSON format.
Or you can execute: 

\begin{command}
python Soccer.py -data=inst.dzn -parser=Soccer_ParserZ.py
\end{command}
where `inst.dzn' is a data Zinc file.

\subsection{Social Golfers}

This is Problem \href{https://www.csplib.org/Problems/prob010/}{010} on CSPLib, and called the Social Golfers Problem.

\paragraph{Description.}
From Warwick Harvey on CSPLib: ``The coordinator of a local golf club has come to you with the following problem.
  In their club, there are 32 social golfers, each of whom play golf once a week, and always in groups of 4.
  They would like you to come up with a schedule of play for these golfers, to last as many weeks as possible, such that no golfer plays in the same group as any other golfer on more than one occasion.
The problem can easily be generalized to that of scheduling $m$ groups of $n$ golfers over $p$ weeks, such that no golfer plays in the same group as any other golfer twice (i.e. maximum socialisation is achieved).''

\paragraph{Data.}

Three integers are required to specify a specific instance. Values of $(m,n,p)$ used for generating the instances in the competition are:
\begin{quote}
(5,5,6), (6,6,7), (7,7,8), (8,8,9), (9,9,10), (8,4,7), (8,4,8), (8,4,9), (8,4,10), (8,4,11)
\end{quote}

\paragraph{Model.}

The two variants (a main one, and another one called ``cnt'') of the \p3 model, in a file `SocialGolfers.py', used for the \x3 competition are:

\begin{boxpy}\begin{python}
@\imp@

nGroups, size, nWeeks = data  # size of the groups
nPlayers = nGroups * size

# g[w][p] is the group admitting on week w the player p
g = VarArray(size=[nWeeks, nPlayers], dom=range(nGroups))

if not variant():
   satisfy(
      # ensuring that two players don't meet more than one time
      [
         If(
            g[w1][p1] == g[w1][p2],
            Then=g[w2][p1] != g[w2][p2]
         ) for w1, w2 in combinations(nWeeks, 2) for p1, p2 in combinations(nPlayers, 2)
      ]
   )

elif variant("cnt"):
   satisfy(
      # ensuring that two players don't meet more than one time
      [
         Sum(g[w][p1] == g[w][p2] for w in range(nWeeks)) <= 1
           for p1, p2 in combinations(nPlayers, 2)
      ]
   )

satisfy(
   # respecting the size of the groups
   [
      Cardinality(
         within=g[w],
         occurrences={i: size for i in range(nGroups)}
      ) for w in range(nWeeks)
   ],
  
   # tag(symmetry-breaking)
   [
      LexIncreasing(g, matrix=True),
    
      [g[0][p] == p // size for p in range(nPlayers)],   
      [g[w][k] == k for k in range(size) for w in range(1, nWeeks)]
   ]
)

if nGroups == size and nGroups + 1 == nWeeks:
   satisfy(
      # tag(redundant)
      [
         [AllDifferent(g[1:, p*size:p*size + size], matrix=True) for p in range(1, nGroups)],
         [g[1][k] == k % size for k in range(size, nPlayers)]
      ]
   )

else:
   satisfy(
      # tag(redundant)
      [
         AllDifferent(g[w][p * size:p * size + size])
           for w in range(1, nWeeks) for p in range(1, nGroups)
      ]
   )
\end{python}\end{boxpy}

This model involves 1 array of variables and global constraints of type \gb{AllDifferent}, \gb{Cardinality}, \gb{Lex}, and \gb{Sum}.
A series of $2*9$ instances has been selected for the competition.
For generating an \x3 instance (file), you can execute for example:
\begin{command}
  python SocialGolfers.py -data=[5,5,6]
  python SocialGolfers.py -data=[5,5,6] -variant=cnt
\end{command}

\subsection{Solitaire Pattern}

\paragraph{Description.}

This is a variant of Peg Solitaire \cite{JMMT_modelling} where a goal state (conﬁguration) with a number of pegs in some speciﬁc arrangement must be reached.
The initial state is the same as that of central Solitaire (i.e., missing peg in the middle of the board).

\vspace{-0.2cm}
\paragraph{Data.}

Three integers are required to specify a specific instance: the position ($ox$,$oy$) of the mising peg, here, always (3,3), and the pattern number $k$. Values of $(ox,oy,k)$ used for generating the instances in the competition are:
\begin{quote}
(3,3,0), (3,3,1), (3,3,2), (3,3,3), (3,3,4), (3,3,5), (3,3,6), (3,3,7), (3,3,8), (3,3,9)
\end{quote}

\paragraph{Model.}

The \p3 model, in `SolitairePattern.py', used for the \x3 competition is:

\begin{boxpy}\begin{python}
@\imp@
from PegSolitaire_Generator import generate_boards

patterns = [  # 10 patterns from Jefferson et al.'s paper
   [(0, 2), (0, 3), (1, 2), (1, 4), (2, 2), (2, 4), (3, 2), (3, 4), (4, 2), (4, 4), (5, 2), (5, 4), (6, 2), (6, 3)],
   [(0, 2), (1, 2), (2, 3), (2, 5), (2, 6), (3, 2), (3, 4), (4, 0), (4, 1), (4, 3), (5, 4), (6, 4)],
   [(0, 2), (0, 3), (0, 4), (1, 2), (2, 0), (2, 3), (2, 5), (2, 6), (3, 0), (3, 2), (3, 4), (3, 6), (4, 0), (4, 1), (4, 3), (4, 6), (5, 4), (6, 2), (6, 3), (6, 4)],
   [(0, 2), (0, 3), (0, 4), (1, 3), (1, 4), (2, 0), (2, 1), (2, 6), (3, 0), (3, 1), (3, 5), (3, 6), (4, 0), (4, 5), (4, 6), (5, 2), (5, 3), (6, 2), (6, 3), (6, 4)],
   [(0, 3), (0, 4), (2, 0), (2, 2), (2, 3), (2, 4), (3, 0), (3, 2), (3, 4), (3, 6), (4, 2), (4, 3), (4, 4), (4, 6), (6, 2), (6, 3)],
   [(0, 2), (0, 4), (1, 4), (2, 0), (2, 1), (2, 3), (2, 6), (3, 2), (3, 4), (4, 0), (4, 3), (4, 5), (4, 6), (5, 2), (6, 2), (6, 4)],
   [(0, 2), (0, 4), (1, 2), (1, 4), (2, 0), (2, 1), (2, 5), (2, 6), (3, 3), (4, 0), (4, 1), (4, 5), (4, 6), (5, 2), (5, 4), (6, 2), (6, 4)],
   [(0, 2), (0, 3), (0, 4), (2, 0), (2, 6), (3, 0), (3, 3), (3, 6), (4, 0), (4, 6), (6, 2), (6, 3), (6, 4)],
   [(0, 3), (3, 0), (3, 3), (3, 6), (6, 3)],
   [(2, 2), (2, 4), (4, 2), (4, 4)]
]

origin_x, origin_y, pattern = data
init_board, final_board = generate_boards("english", origin_x, origin_y)
n, m = len(init_board), len(init_board[0])

points = [(i, j) for i in range(n) for j in range(m) if init_board[i][j] is not None]
for i, j in points:
   final_board[i][j] = 1 if (i, j) in patterns[pattern] else 0

horizon = sum(sum(v for v in row if v) for row in init_board)
           - sum(sum(v for v in row if v) for row in final_board)
nMoves = horizon

_m = [[
         (i, j, i + 1, j, i + 2, j),
         (i, j, i, j + 1, i, j + 2),
         (i, j, i - 1, j, i - 2, j),
         (i, j, i, j - 1, i, j - 2)
       ] for i, j in points ]

TR = sorted((t[0:2], t[2:4], t[4:6]) for row in _m for t in row
               if 0 <= t[4] < n and 0 <= t[5] < m and init_board[t[4]][t[5]] is not None)
nTransitions = len(TR)

def independent(tr1, tr2):
   return len(set(tr1 + tr2)) == 6

def unchanged(i, j, t):
   V = [k for k, (p1, p2, p3) in enumerate(TR) if (i, j) in (p1, p2, p3)]
   cond = conjunction(y[t] != k for k in V)
   return cond == (x[t][i][j] == x[t + 1][i][j])

def to0(i, j, t):
   V = [k for k, (p1, p2, p3) in enumerate(TR) if (i, j) in (p1, p2)]
   cond = disjunction(y[t] == k for k in V)
   return cond == both(x[t][i][j] == 1, x[t + 1][i][j] == 0)

def to1(i, j, t):
   V = [k for k, (p1, p2, p3) in enumerate(TR) if (i, j) == p3]
   cond = disjunction(y[t] == k for k in V)
   return cond == both(x[t][i][j] == 0, x[t + 1][i][j] == 1)
        
T2 = [(k, q) for k, tr1 in enumerate(TR) for q, tr2 in enumerate(TR) if k != q and (k < q or not independent(tr1,tr2)) and len({tr1[0],tr1[1],tr2[0],tr2[1]})==4 and tr1[2]!=tr2[2]]

T3 = [(k, q) for k, (p1, p2, p3) in enumerate(TR) for q, (q1, q2, q3) in enumerate(TR)
      if k == q or (p1 == q2 and p2 == q1)]

# x[t][i](j] is the value at row i and column j at time t
x = VarArray(size=[nMoves + 1, n, m], dom=lambda t, i, j: None if init_board[i][j] is None else {0, 1})

# y[t] is the move (transition) performed at time t
y = VarArray(size=nMoves, dom=range(nTransitions))

satisfy(
   # setting the initial board
   x[0] == init_board,

   # setting the final board
   x[-1] == final_board,
   
   # tag(symmetry-breaking)
   [(y[i], y[i + 1]) in T2 for i in range(nMoves - 1)],
   
   # tag(redundant-constraints)
   [[(y[i], y[i + 2]) not in T3 for i in range(nMoves - 2)]],

   # handling unchanged situations
   [unchanged(i, j, t) for (i, j) in points for t in range(nMoves)],
   
   # handling situations where a peg disappears
   [to0(i, j, t) for (i, j) in points for t in range(nMoves)],

   # handling situations where a peg appears
   [to1(i, j, t) for (i, j) in points for t in range(nMoves)]
   
   # tag(redundant-constraints)
   [Sum(x[t]) == 32 - t for t in range(nMoves)],
)
\end{python}\end{boxpy}

This model involves 2 arrays of variables and global constraints of type \gb{Sum} and \gb{Table}.
A series of 10 instances has been selected for the competition.
For generating an \x3 instance (file), you can execute for example:
\begin{command}
  python SolitairePattern.py -data=[3,3,0]
\end{command}

\subsection{Subisomorphism}

This problem has already been selected in previous XCSP competitions.
This satisfaction problem only involves (ordinary) table constraints and the global constraint \gb{AllDifferent}.
A series of 11 instances has been selected for the competition.

\subsection{Takuzu}

\paragraph{Description.}

From \href{https://en.wikipedia.org/wiki/Takuzu}{Wikipedia}: ``Takuzu, also known as Binairo, is a logic puzzle involving placement of two symbols, often 1s and 0s, on a rectangular grid.
The objective is to fill the grid with 1s and 0s, where there is an equal number of 1s and 0s in each row and column and no more than two of either number adjacent to each other.
Additionally, there can be no identical rows or columns.''
\paragraph{Data.}

In our model, only one integer is required to specify a specific instance: the value of $n$ (order of the grid).
In the model, it is possible to specify some clues (but, in our case, we don't have any, and we indicate this with value `None').  
The values used for generating the 2024 competition instances are:
\begin{quote}
30, 40, 50, 60, 70, 80, 90, 100, 110, 120, 150 200
\end{quote}

\paragraph{Model.}
The \p3 model, in a file `Takuzu.py', used for the competition is: 

\begin{boxpy}\begin{python}
@\imp@

n, grid = data
m = n // 2

# x[i][j] is the value in the cell of the grid at coordinates (i,j)
x = VarArray(size=[n, n], dom={0, 1})

satisfy(
   # ensuring that each row has the same number of 0s and 1s
   [Sum(x[i]) == m for i in range(n)],

   # ensuring that each colum has the same number of 0s and 1s
   [Sum(x[:, j]) == m for j in range(n)],

   # ensuring no more than two adjacent equal values on each row
   [
      either(
         x[i][j] != x[i - 1][j],
         x[i][j] != x[i + 1][j]
      ) for i in range(1, n - 1) for j in range(n)
   ],
   
   # ensuring no more than two adjacent equal values on each column
   [
      either(
         x[i][j] != x[i][j - 1],
         x[i][j] != x[i][j + 1]
      ) for j in range(1, n - 1) for i in range(n)
   ],

   # forbidding identical rows
   AllDifferentList(x[i] for i in range(n)),
   
   # forbidding identical columns
   AllDifferentList(x[:, j] for j in range(n)),
)
\end{python}\end{boxpy}

This model involves a two-dimensional array of variables $x$, and global constraints of type \gb{AllDifferentList} and \gb{Sum}.
A series of 12 instances has been selected for the competition.
For generating an \x3 instance (file), you can execute for example:
\begin{command}
python Takuzu.py -data=[10,None]
\end{command}

\subsection{WordSquare}

\paragraph{Description.}

From \href{https://en.wikipedia.org/wiki/Word_square}{Wikipedia}: ``A word square is a type of acrostic.
It consists of a set of words written out in a square grid, such that the same words can be read both horizontally and vertically.
The number of words, which is equal to the number of letters in each word, is known as the order of the square.''

\paragraph{Data.}

In our model, one integer $n$ (order of the square) as well a dictionary file are required to specify a specific instance.
The values of $n$ used for generating the 2024 competition instances are:
\begin{quote}
7, 8, 9, 10, 11
\end{quote}

\paragraph{Model.}
The three variants (called ``hak'', ``tab1'', ``tab2'') of the \p3 model, in a file `WordQuare.py', used for the \x3 competition are:

\begin{boxpy}\begin{python}
@\imp@

n, dict_name = data

words = []
for line in open(dict_name):
   code = alphabet_positions(line.strip().lower())
   if len(code) == n:
      words.append(code)
words, nWords = cp_array(words), len(words)

if variant("hak"):

   # y[k] is the word chosen for the kth position (row and column)
   y = VarArray(size=n, dom=range(nWords))

   satisfy(
      # ensuring coherence of words
      [words[y[i], j] == words[y[j], i] for i in range(n) for j in range(n)],

      # ensuring different words
      AllDifferent(y)
    )

elif variant("tab1"):
      
   #  x[i][j] is the letter, number from 0 to 25, at row i and column j
   x = VarArray(size=[n, n], dom=range(26))

   # y[k] is the word chosen for the kth position (row and column)
   y = VarArray(size=n, dom=range(nWords))

   satisfy(
      [(y[i], x[i]) in enumerate(words) for i in range(n)],

      [x[i] == x[:, i] for i in range(n)],

      AllDifferent(y)
   )

elif variant("tab2"):

   # x[i][j] is the letter, number from 0 to 25, at row i and column j
   x = VarArray(size=[n, n], dom=range(26))

   satisfy(
      [x[i] in words for i in range(n)],

      [x[i] == x[:, i] for i in range(n)],

      AllDifferentList(x[i] for i in range(n))
   )
\end{python}\end{boxpy}

Note that ``hak'' corresponds to the model proposed by Hakan Kjellerstrand (see \href{http://www.hakank.org/minizinc/word_square.mzn}{it}).
A series of $3*5$ instances has been selected for the competition.
For generating an \x3 instance (file), you can execute for example:
\begin{command}
python WordSquare.py -data=[8,ogd2008] -variant='hak'
python WordSquare.py -data=[8,ogd2008] -variant='tab1'
python WordSquare.py -data=[8,ogd2008] -variant='tab2'
\end{command}

wher `ogd2008' is a dictionary file.

\section{COP}

\subsection{Aircraft Assembly Line}

\paragraph{Description.}

This problem has been proposed by Stéphanie Roussel from ONERA (Toulouse), and comes from an aircraft manufacturer.
The objective is to schedule tasks on an aircraft assembly line in order to minimize the overall number of operators required on the line.
The schedule must satisfy several operational constraints, the main ones being:
\begin{itemize}
\item tasks are assigned on a unique workstation (on which specific machines are available);
\item the takt-time, i.e., the duration during which the aircraft stays on each workstation, must be respected;
\item capacity of aircraft zones in which operators perform the tasks must never be exceeded;
\item zones can be neutralized by some tasks, i.e., it is not possible to work in those zones during the tasks execution.
\end{itemize}

Note that similar problems have been studied in \cite{RCP23}, where the authors are interested in the design of assembly lines (with similar instances).

\paragraph{Data.}
As an illustration of data specifying an instance of this problem, we have:

\begin{json}
{
  "takt": 1440,
  "nTasks": 199,
  "nMachines": 5,
  "nAreas": 48,
  "areasCapacities": [1,1,..,1],
  "tasksPerMachine": [
    [49,50,51,52,53],
    ...,
    [197]
  ],
  "nMaxOpsPerStation": [10,10,10,10],
  "neutralizedAreas": [
    [12,15,24,28,31,37,39,46],
    ...,
    []
  ],
  "operators": [0,1,...,1],
  "tasksPerAreas": [
    [2,5,11,13],
    ...,
    [6,7,165,166,167]
  ],
  "usedAreas":[
    [0,0,...,0],
    ...,
    [0,0,...,0]
  ],
  "durations": [0,109,...,470],
  "precedences": [
    [8,9],
    ...,
    [198,190]
  ],
  "nStations": 4,
  "machines":[
    [0,0,0,0,0],
    ...,
    [0,0,1,1,1]
  ]
}
\end{json}

\paragraph{Model.}
\bigskip
The \p3 model, in a file `AircraftAssemblyLine.py', used for the competition is:

\begin{boxpy}\begin{python}
@\imp@

takt, areas, stations, tasks, tasksPerMachine, precedences = data
nAreas, nStations, nTasks = len(areas), len(stations), len(tasks)
nMachines = len(tasksPerMachine)

areaCapacities, areaTasks = zip(*areas)  # nb of operators who can work, and tasks per area
stationMachines, stationMaxOperators = zip(*stations)
durations, operators, usedAreaRooms, neutralizedAreas = zip(*tasks)
usedAreas = [set(j for j in range(nAreas) if usedAreaRooms[i][j] > 0) for i in range(nTasks)]

def station_of_task(i):
   r = next((j for j in range(nMachines) if i in tasksPerMachine[j]), -1)
   return -1 if r == -1 else next(j for j in range(nStations) if stationMachines[j][r] == 1)

stationOfTasks = [station_of_task(i) for i in range(nTasks)]  # -1 if can be everywhere

# x[i] is the starting time of the ith task
x = VarArray(size=nTasks, dom=range(takt * nStations + 1))

# z[j] is the number of operators at the jth station
z = VarArray(size=nStations, dom=lambda i: range(stationMaxOperators[i] + 1))

satisfy(
   # respecting the final deadline
   [x[i] + durations[i] <= takt * nStations for i in range(nTasks)],

   # ensuring that tasks start and finish in the same station
   [x[i] // takt == (x[i] + max(0, durations[i] - 1)) // takt
      for i in range(nTasks) if durations[i] != 0],

   # ensuring that tasks are put on the right stations (wrt needed machines)
   [x[i] // takt == stationOfTasks[i] for i in range(nTasks) if stationOfTasks[i] != -1],

   # respecting precedence relations
   [x[i] + durations[i] <= x[j] for (i, j) in precedences],

   # respecting limit capacities of areas
   [
      Cumulative(
         Task(origin=x[t], length=durations[t], height=usedAreaRooms[t][i])
            for t in areaTasks[i]
      ) <= areaCapacities[i] for i in range(nAreas) if len(areaTasks[i]) > 1
   ],

   # computing/restricting the number of operators at each station
   [
      Cumulative(
         Task(origin=x[t], length=durations[t], height=operators[t] * (x[t] // takt == j))
            for t in range(nTasks)
      ) <= z[j] for j in range(nStations)
   ],
    
   # no overlapping between some tasks 
   [NoOverlap(tasks=[(x[i], durations[i]), (x[j], durations[j])])
      for i in range(nTasks) for j in range(nTasks)
         if i != j and len(usedAreas[i].intersection(neutralizedAreas[j])) > 0],

   # avoiding tasks using the same machine to overlap
   [NoOverlap(tasks=[(x[j], durations[j]) for j in tasksPerMachine[i]])
      for i in range(nMachines) if len(tasksPerMachine[i]) > 1]
)

minimize(
   # minimizing the number of operators
   Sum(z)
)
\end{python}\end{boxpy} 

This involves 2 arrays of variables and 3 types of constraints: \gb{Intension}, \gb{Cumulative} and \gb{NoOverlap}.
A series of 20 instances has been selected from data files generated by Stéphanie Roussel from ONERA (Toulouse).

For generating an \x3 instance (file), you can execute for example:
\begin{command}
python AircraftAssemblyLine.py -data=1-178.json -parser=Aircraft_Converter.py
\end{command}

where `1-178.json' is a data file in JSON format, and `Aircraft\_Converter.py' is a converter tool allowing us to pass from one JSON format to another that is easier to handle (with respect to the model).
See how the initial 15 data fields have been reduced to 6 data fields only. 
Note that for saving data in JSON files, you can add the option `-export' (or `-dataexport').

\subsection{AztecDiamondSym}

\paragraph{Description.}

This is a variant of Aztec Diamond, turned into an optimization problem.
See \href{https://en.wikipedia.org/wiki/Aztec_diamond}{Wikipedia}.

\paragraph{Data.}

The values of $n$ used for generating the 2024 competition instances are:
\begin{quote}
3, 4 5, 6, 7, 8, 9, 10, 12, 15
\end{quote}

\paragraph{Model.}
The \p3 model, in a file `AztecDimaondSym.py', used for the competition is: 

\begin{boxpy}\begin{python}
@\imp@

n = data

def n_dominos(k):
   return 6 if k == 2 else n_dominos(k - 1) + 2 * k

nDominos = n_dominos(n)

T3 = [(k, k, ANY) for k in range(nDominos)]
     + [(k, ANY, k) for k in range(nDominos)]

T5 = [(k, k, ANY, ANY, ANY) for k in range(nDominos)]
     + [(k, ANY, k, ANY, ANY) for k in range(nDominos)]
     + [(k, ANY, ANY, k, ANY) for k in range(nDominos)]
     + [(k, ANY, ANY, ANY, k) for k in range(nDominos)]

def valid(i, j):
   if i < 0 or i >= n * 2 or j < 0 or j >= n * 2:
      return False
   if i < n - 1 and (j < n - 1 - i or j > n + i):
      return False
   if i > n and (j < i - n or j > 3 * n - i - 1):
      return False
   return True

def top_left(i, j):
   return valid(i, j) and not valid(i, j - 1) and not valid(i - 1, j)

def top_right(i, j):
   return valid(i, j) and not valid(i, j + 1) and not valid(i - 1, j)

def bot_left(i, j):
   return valid(i, j) and not valid(i, j - 1) and not valid(i + 1, j)

def bot_right(i, j):
   return valid(i, j) and not valid(i, j + 1) and not valid(i + 1, j)

# x[i][j] is the number of the domino in cell (i,j)
x = VarArray(size=[2 * n, 2 * n], dom=lambda i, j: range(nDominos) if valid(i, j) else None)

m = [[x[i][j]] + [x[k][q] for k, q in [(i, j - 1), (i, j + 1), (i - 1, j), (i + 1, j)]
      if valid(k, q)] for i in range(2 * n) for j in range(2 * n) if valid(i, j)]

scp3, scp5 = [t for t in m if len(t) == 3], [t for t in m if len(t) == 5]

satisfy(

   # ensuring valid positioning of dominos (over sequences of length 3)
   [scp in T3 for scp in scp3],

   # ensuring valid positioning of dominos (over sequences of length 5)
   [scp in T5 for scp in scp5],

   # ensuring the two pieces of each domino occurs twice
   Cardinality(x, occurrences={k: 2 for k in range(nDominos)})
)

minimize(
   Sum(
      x[i][j] * (abs(n-i) * abs(n-j)) for i in range(2*n) for j in range(2*n) if valid(i,j)
   )
)
\end{python}\end{boxpy}

This model involves a two-dimensional array of variables $x$, and (global) constraints of type \gb{Cardinality}, \gb{Sum} and \gb{Table}.
A series of 10 instances has been selected for the competition.
For generating an \x3 instance (file), you can execute for example:
\begin{command}
python AztecDiamondSym.py -data=10
\end{command}

\subsection{Bin Packing}

\paragraph{Description.}

The bin packing problem (BPP) can be informally defined in a very simple way.
We are given $n$ items, each having an integer weight $w_j$ ($j = 1, \dots, n$), and an unlimited number of identical bins of integer capacity $c$.
The objective is to pack all the items into the minimum number of bins so that the total weight packed in any bin does not exceed the capacity $c$.
Note that many resources can be found at \href{https://site.unibo.it/operations-research/en/research/bpplib-a-bin-packing-problem-library}{BPPLIB} – A Bin Packing Problem Library.

\paragraph{Data.}
As an illustration of data specifying an instance of this problem, we have:

\begin{json}
{
  "binCapacity": 100,
  "itemWeights": [30, 31, ..., 99]
}
\end{json}

\paragraph{Model.}
\bigskip
Two variants (a main one, and a second one called ``table'') of a first \p3 model, in a file `BinPacking.py', have been used for the competition: 
       
\begin{boxpy}\begin{python}
@\imp@
from itertools import groupby
from math import ceil

capacity, weights = data  # bin capacity and item weights
weights.sort()  # in case weights are not sorted
nItems = len(weights)

def n_bins():
   cnt = 0
   curr_load = 0
   for i, weight in enumerate(weights):
      curr_load += weight
      if curr_load > capacity:
         cnt += 1
         curr_load = weight
   return cnt

def max_items_per_bin():
   curr = 0
   for i, weight in enumerate(weights):
      curr += weight
      if curr > capacity:
         return i
   return -1

def w(a, b, *, bar=False):
   if bar:
      return [i for i, weight in enumerate(weights) if a <= weight <= b]
   return [i for i, weight in enumerate(weights) if a < weight <= b]

def lb2(v=None):
   half = len(w(capacity // 2, capacity))
   if v is None:
      return max(lb2(vv) for vv in range(capacity // 2 + 1))
   return half + max(0, ceil(sum(weights[i] for i in w(v, capacity - v, bar=True)) / capacity - len(w(capacity // 2, capacity - v))))

nBins, maxPerBin = n_bins(), max_items_per_bin()

# x[i][j] is the weight of the jth object put in the ith bin (or 0)
x = VarArray(size=[nBins, maxPerBin], dom={0, *weights})

# z is the number of used bins
z = Var(range(lb2(), nBins + 1))

if not variant():
   satisfy(
      # not exceeding the capacity of each bin
      [Sum(x[i]) <= capacity for i in range(nBins)],

      # items are stored decreasingly in each bin according to their weights
      [Decreasing(x[i]) for i in range(nBins)]
   )

elif variant("table"):
   def table():
      def table_recursive(n_stored, i, curr):
         tmp[n_stored] = weights[i]
         curr += weights[i]
         tuples.append(tuple(tmp[j] if j < n_stored + 1 else 0 for j in range(maxPerBin)))
         for j in range(i):
            if curr + weights[j] > capacity:
               break
            if j == i - 1 or weights[j] != weights[j + 1]:
               table_recursive(n_stored + 1, j, curr)

      tmp = [0] * maxPerBin
      tuples = [tuple(tmp)]
      for i in range(nItems):
         if i == nItems - 1 or weights[i] != weights[i + 1]:
            table_recursive(0, i, 0)
      return tuples

   T = table()
   satisfy(
      x[i] in T for i in range(nBins)
   )

satisfy(
   # computing the number of used bins
   z == Sum(x[i][0] != 0 for i in range(nBins)),

   # ensuring that each item is stored in a bin
   Cardinality(
      within=x,
      occurrences={0: nBins * maxPerBin - nItems}
                    | {wgt: len(list(t)) for wgt, t in groupby(weights)}
   ),

   # tag(symmetry-breaking)
   LexDecreasing(x)
)

minimize(
   # minimizing the number of used bins
   z  
)
\end{python}\end{boxpy} 

This model involves one array of variables $x$, a stand-alone variable $z$ and (global) constraints of type \gb{Cardinality}, \gb{Lex}, \gb{Sum} and \gb{Table}.
A series of $6+3$ instances has been selected from the two variants of this first model.

For generating an \x3 instance (file), you can execute for example:
\begin{command}
  python BinPacking.py -data=inst.json
  python BinPacking.py -data=inst.json -variant=table
\end{command}

where `inst.json' is a data file in JSON format.

\bigskip
The second \p3 model, in a file `BinPacking2.py', used for the competition is: 
       
\begin{boxpy}\begin{python}
@\imp@
from math import ceil

capacity, weights = data  # bin capacity and item weights
weights.sort()
nItems = len(weights)

def n_bins():
   cnt = 0
   curr_load = 0
   for weight in weights:
      curr_load += weight
      if curr_load > capacity:
         cnt += 1
         curr_load = weight
   return cnt

def series():
   t = []
   start = 0
   for i, weight in enumerate(weights):
      if weight == weights[start]:
         continue
      if start < i - 1:
         t.append((start, i - start))
      start = i
   return t

def w(a, b, *, bar=False):
   if bar:
      return [i for i, weight in enumerate(weights) if a <= weight <= b]
   return [i for i, weight in enumerate(weights) if a < weight <= b]

def lb2(v=None):
   half = len(w(capacity // 2, capacity))
   if v is None:
      return max(lb2(vv) for vv in range(capacity // 2 + 1))
   return half + max(0, ceil(sum(weights[i] for i in w(v, capacity - v, bar=True)) / capacity - len(w(capacity // 2, capacity - v))))

nBins = n_bins()
n_exceeding = len([weight for weight in weights if weight > capacity // 2])

# x[i] is the bin (number) where is put the ith item
x = VarArray(size=nItems, dom=range(nBins))

# z is the number of used bins
z = Var(range(lb2(), nBins + 1))

satisfy(
   # ensuring that the capacity of each bin is not exceeded
   BinPacking(x, sizes=weights) <= capacity,

   # ensuring a minimum number of bins
   z == NValues(x),  # >= ceil(sum(weights) / capacity),

   # tag(symmetry-breaking)
   [Increasing(x[s:s + l]) for (s, l) in series()],

   # tag(symmetry-breaking)
   [x[nItems - n_exceeding + i] == i for i in range(n_exceeding)]
)

minimize(
   # minimizing the number of used bins
   z  # NValues(x)
)
\end{python}\end{boxpy}

This model involves one array of variables $x$, a stand-alone variable $z$ and (global) constraints of type \gb{BinPacking}, and \gb{NValues}.
A series of $6$ instances has been selected for this second model.

For generating an \x3 instance (file), you can execute for example:
\begin{command}
  python BinPacking2.py -data=inst.json
\end{command}

where `inst.json' is a data file in JSON format.

\subsection{Cargo}

\paragraph{Description.}

This problem is described in \cite{SS_cargo,BBSS_local}.
This is a real-world cargo assembly planning problem arising in a coal supply chain.
The cargoes are built on the stockyard at a port terminal from coal delivered by trains.
Then the cargoes are loaded onto vessels.
Only a limited number of arriving vessels is known in advance.
The goal is to minimize the average delay time of the vessels over a long planning period.

\paragraph{Data.}

As an illustration of data specifying an instance of this problem, we have:
\begin{json}
{
  "H": 1700,
  "T": 42000,
  "stackBefore": 30,
  "limits": {
    "dailyStacking": 950,
    "nReclaimers": 2,
    "maxReclaimingGap": 300,
    "maxDelay": 14200,
    "sumMaxDelay": 330000
  },
  "factors": {
    "meter": 1,
    "time": 1440,
    "tonnage": 1000,
    "length": 17,
    "hour": 60
  },
  "etas": [14189, 15462, ..., 32993],
  "piles": [
    {"whichV": 0, "dS_": 3, "dR": 545},
    {"whichV": 1, "dS_": 7, "dR": 561},
    ...
    {"whichV": 13, "dS_": 5, "dR": 821}
  ]
}
\end{json}

\paragraph{Model.}

An original model was proposed in the context of Minizinc challenges.
The \p3 model, in a file `Cargo.py', used for the \x3 competition is close to (can be seen as the close translation of) it: 

\begin{boxpy}\begin{python}
@\imp@

H, T, stackBefore, limits, coeffs, eta, piles = data
vessels, r_sd, rd = zip(*piles)  # r_sd is for rounded stacking duration
nVessels, nPiles = len(eta), len(piles)

V, P = range(nVessels), range(nPiles)

vrd = [sum(rd[i] for i in P if vessels[i] == j) for j in V]  # summed reclaiming durations
dailyStackingTonnages = [(rd[i] * coeffs.tonnage) // (r_sd[i] * coeffs.time) for i in P]
lengths = [((rd[i]*coeffs.length) // coeffs.hour+coeffs.meter-1) // coeffs.meter for i in P]
lastPiles = [max(i for i in P if vessels[i] == j) for j in V]  

# r_st[i] is the rounded stacking start time of the ith stockpile
r_st = VarArray(size=nPiles, dom=range(T // coeffs.time + 1))

# st[i] is the precise stacking start time of the ith stockpile
st = VarArray(size=nPiles, dom=range(T + 1))

# y[i] is the rounded position of the ith stockpile
y = VarArray(size=nPiles, dom=range(H // coeffs.meter + 1))

# rt[i] is the reclaiming start time of the ith stockpile
rt = VarArray(size=nPiles, dom=range(T + 1))

# r_dt[i] is the rounded total duration time of processing the ith stockpile
r_dt = VarArray(size=nPiles, dom=range(T // coeffs.time + 1))

# dt[i] is the precise total duration time of processing the ith stockpile
dt = VarArray(size=nPiles, dom=range(T + 1))

# ft[i] is the finishing time, when the ith vessel is ready to leave
ft = VarArray(size=nVessels, dom=range(T + 1))

satisfy(
   # linking precise and rounded stacking start times
   [st[i] == r_st[i] * coeffs.time for i in P],

   # linking precise and rounded total processing times
   [dt[i] == r_dt[i] * coeffs.time for i in P],
   
   # making vessels ready when reclaiming is finished
   [ft[j] == rt[lastPiles[j]] + rd[lastPiles[j]] for j in V],
   
   # computing total processing times
   [dt[i] >= rt[i] + rd[i] - st[i] for i in P],
   
   # finishing in time
   [r_st[lastPiles[j]] + r_dt[lastPiles[j]] <= (T+coeffs.time-1) // coeffs.time for j in V],
   
   # reclaiming of a stockpile cannot start before the ETA of its vessel
   [rt[i] >= eta[vessels[i]] for i in P],
   
   # stacking of a stockpile starts when possible
   [st[i] >= eta[vessels[i]] - stackBefore * coeffs.time for i in P],
   
   # respecting the continuous reclaim time limit (e.g., 5 hours)
   [rt[i + 1] <= rt[i] + rd[i] + limits.maxReclaimingGap for i in range(nPiles - 1)
       if vessels[i] == vessels[i + 1]],
   
   # stacking of a stockpile has to complete before reclaiming can start
   [r_st[i] + r_sd[i] <= rt[i] // coeffs.time for i in P],
   
   # ensuring stockpiles fit on their pads
   [y[i] + lengths[i] <= H // coeffs.meter for i in P],
   
   # respecting the reclaim order of the stockpiles of a vessel
   [rt[i] + rd[i] <= rt[i + 1] for i in range(nPiles - 1) if vessels[i] == vessels[i + 1]],
   
   # computing the sum of vessel delays
   Sum(ft[j] - eta[j] - vrd[j] for j in V) <= limits.sumMaxDelay,
   
   # not exceeding a maximum delay
   [ft[j] - eta[j] - vrd[j] <= limits.maxDelay for j in V],
   
   # not overlapping stockpiles in space and time
   NoOverlap(
      origins=(r_st, y),
      lengths=(r_dt, lengths)  
   ),
   
   # not exceeding the pad lengths
   Cumulative(
      origins=r_st,
      lengths=r_dt,
      heights=lengths
   ) <= H // coeffs.meter,
   
   # not exceeding the daily stacking capacity
   Cumulative(
      origins=r_st,
      lengths=r_sd,
      heights=dailyStackingTonnages
   ) <= limits.dailyStacking,
   
   # ensuring that there are always enough reclaimers
   Cumulative(
      origins=rt,
      lengths=rd,
      heights=1
   ) <= limits.nReclaimers
)

minimize(
   Sum(ft[j] - eta[j] - vrd[j] for j in range(4, nVessels - 5))
)
\end{python}\end{boxpy}

This model involves seven arrays of variables, and (global) constraints of type \gb{Cumulative}, \gb{Element}, \gb{NoOverlap} and \gb{Sum}.

A series of 16 instances has been selected for the competition (data files come from Minizinc challenges).
For generating an \x3 instance (file), you can execute for example:
\begin{command}
python Cargo.py -data=<datafile.json>
python Cargo.py -data=<datafile.dzn> -parser=Cargo_ParserZ.py
\end{command}

\subsection{Charlotte}

\paragraph{Description.}

Charlotte is a freelance nurse who works in shifts with two colleagues.
Together, they work shifts which they share out over long periods (several weeks).

\paragraph{Data.}

Two integers are required to specify a specific instance: the number of weeks ($n$) and a seed ($s$) for generating randomly unwished work days. Values of $(n,s)$ used for generating the instances in the competition are:
\begin{quote}
(6,0), (6,1), (6,2), (12,0), (12,1), (12,2), (18,0), (18,1), (18,2), (24,0), (24,1), (24,2)
\end{quote}

\paragraph{Model.}

The \p3 model, in a file `Charlotte.py', used for the \x3 competition is:

\begin{boxpy}\begin{python}
@\imp@
import random

nWeeks, seed = data 
nNurses = 3
assert nWeeks % nNurses == 0
nDays = nWeeks * 7

nShortShifts, nLongShifts = (nWeeks * 5) // nNurses, (nWeeks * 7) // nNurses
nOffShifts = nDays - nShortShifts - nLongShifts
OFF, SHORT, LONG = shifts = range(3)  # off, short day, long day
nShifts = len(shifts)

MAX_WITHOUT_OFF = 7  # so, at least one day off every sequence of 8 days

weekend_days = [i for i in range(nDays) if i % 7 in (5, 6)]
business_days = [i for i in range(nDays) if i not in weekend_days]

random.seed(seed)
unwished = [[random.randint(0, 100) for _ in range(nDays)] for _ in range(nNurses)]

T1 = [(0, 0, ANY), (0, 1, 1), (0, 1, 2), (0, 2, 1), (0, 2, 2), (1, 0, 0), (1, 1, ANY),
       (1, 2, ANY), (2, 0, 0), (2, 1, ANY), (2, 2, ANY)]

T2 = [tuple(0 if i <= ii < i + 5 else ANY for ii in range(21)) for i in range(21 - 4)]

T3 = [tuple(0 if i == ii else ANY for ii in range(MAX_WITHOUT_OFF + 1))
        for i in range(MAX_WITHOUT_OFF + 1)]

# x[i][j] is the shift for the jth nurse on the ith day
x = VarArray(size=[nDays, nNurses], dom=lambda i, j: shifts if i in business_days
                                                           else {OFF, LONG})

# y[j] is the nurse working on the jth weekend (and its multiple)
y = VarArray(size=nNurses, dom=range(nNurses))

satisfy(
  # ensuring different shifts on business days
  [AllDifferent(x[i]) for i in business_days]

  # only one nurse working on weekend days
  [Count(x[i], value=OFF) == 2 for i in weekend_days],
  
  # ensuring equity for off days
  [Sum(x[i][j] == OFF for i in range(nDays)) == nOffShifts for j in range(nNurses)],
  
  # ensuring equity for small days
  [Sum(x[i][j] == SHORT for i in range(nDays)) == nShortShifts for j in range(nNurses)],
  
  # ensuring equity for long days
  [Sum(x[i][j] == LONG for i in range(nDays)) == nLongShifts for j in range(nNurses)],
  
  # ensuring a permutation (for ordering worked weekends)
  AllDifferent(y),
  
  # ensuring the same nurse working a long day on friday, saturday and sunday
  [
     [x[k * 7 + 4][y[k % 3]] == LONG for k in range(nWeeks)],
     [x[k * 7 + 5][y[k % 3]] == LONG for k in range(nWeeks)],
     [x[k * 7 + 6][y[k % 3]] == LONG for k in range(nWeeks)]
  ],
                    
  # ensuring at least two days off, and two days working for every sequence of three days
  [x[i:i + 3, j] in T1 for i in range(nDays - 2) for j in range(nNurses)],

  # at least, 5 consecutive days off every three weeks
  [x[k * 7:(k + 3) * 7, j] in T2 for k in range(nWeeks - 2) for j in range(nNurses)],
  
  # at least, one day off at every sequence of 8 days
  [x[i:i + 8, j] in T3 for i in range(nDays - 7) for j in range(nNurses)]
)

minimize(
   Sum(
      x[:, j] * unwished[j] for j in range(nNurses)
   )
)
\end{python}\end{boxpy}

This model involves 2 arrays of variables and (global) constraints of type \gb{AllDifferent}, \gb{Count}, \gb{Sum} and \gb{Table}.

A series of 12 instances has been selected for the competition.
For generating an \x3 instance (file), you can execute for example:
\begin{command}
  python Charlotte.py -data=[12,0]
\end{command}

\subsection{Drinking}

\paragraph{Description.}

From Page 184 in \cite{MS_book}: ``In the drinking game, one must drink one glass everytime a number is reached which is divisible by 7 or divisible by 5,
unless the previous drink was taken less than 8 numbers ago.''
Here, we add an objective function.

\paragraph{Data.}

In our model, only one integer is required to specify a specific instance: the number of minutes $n$.
The values used for generating the 2024 competition instances are:
\begin{quote}
50, 100, 200, 400, 700, 10000, 20000, 50000, 100000, 200000
\end{quote}

\paragraph{Model.}
The \p3 model, in a file `Drinking.py', used for the competition is: 

\begin{boxpy}\begin{python}
@\imp@

n = data  # number of minutes (time slots)

# x[i] is 1 iff time i is drinking time
x = VarArray(size=n, dom={0, 1})

# y[i] is the  number of drinking times the 8 last minutes before time i
y = VarArray(size=n, dom=range(9))  

satisfy(
   # computing the number of drinking times every 8 minutes
   y[t] == Sum(x[max(t - 8, 0):max(t, 1)]) for t in range(1, n)

   # drink when time divisible by 5 or 7 and no drinking in the last 8 minutes
   [(y[t] == 0) == (x[t] == 1) for t in range(1, n) if t % 5 == 0 or t % 7 == 0]
)

minimize(
   Sum(x)
)
\end{python}\end{boxpy}

This model involves two arrays of variables, and (global) constraints of type \gb{Sum}.
A series of 10 instances has been selected for the competition.
For generating an \x3 instance (file), you can execute for example:
\begin{command}
python Drinking.py -data=700
\end{command}

\subsection{Fool Solitaire}

\paragraph{Description.}

From \cite{JMMT_modelling}: ``An optimisation variation of Peg Solitaire (named Fool’s Solitaire by Berlekamp, Conway and Guy \cite{BCG_bwinning}) is to reach a position where no further moves are possible in the shortest sequence of moves.'' 

\paragraph{Data.}

Two integers are required to specify a specific instance: the position ($ox$,$oy$) of the mising peg. Values of $(ox,oy)$ used for generating the instances in the competition are:
\begin{quote}
(0,2), (0,3), (0,4), (1,2), (1,3), (1,4), (2,0), (2,1), (2,2), (2,3), (2,4), (2,5), (2,6), (3,0), (3,1), (3,2), (3,3)
\end{quote}

\paragraph{Model.}

The \p3 model, in a file `FoolSolitaire.py', used for the \x3 competition is:

\begin{boxpy}\begin{python}
@\imp@
from PegSolitaire_Generator import generate_boards

origin_x, origin_y = data

init_board, final_board = generate_boards("english", origin_x, origin_y)
n, m = len(init_board), len(init_board[0])

points = [(i, j) for i in range(n) for j in range(m) if init_board[i][j] is not None]
horizon = sum(sum(v for v in row if v) for row in init_board)
           - sum(sum(v for v in row if v) for row in final_board)
nMoves = horizon

_m = [[
         (i, j, i + 1, j, i + 2, j),
         (i, j, i, j + 1, i, j + 2),
         (i, j, i - 1, j, i - 2, j),
         (i, j, i, j - 1, i, j - 2)
       ] for i, j in points ]

TR = sorted((t[0:2], t[2:4], t[4:6]) for row in _m for t in row
               if 0 <= t[4] < n and 0 <= t[5] < m and init_board[t[4]][t[5]] is not None)
nTransitions = len(TR)
NADA = nTransitions

def independent(tr1, tr2):
    return len(set(tr1 + tr2)) == 6

T2 = [(k, q) for k, tr1 in enumerate(TR) for q, tr2 in enumerate(TR) if k != q and
       (k < q or not independent(tr1, tr2)) and len({tr1[0], tr1[1], tr2[0], tr2[1]}) == 4
       and tr1[2] != tr2[2]] + [(ANY, NADA)]

T3 = [(k, q) for k, (p1, p2, p3) in enumerate(TR) for q, (q1, q2, q3) in enumerate(TR)
       if k == q or (p1 == q2 and p2 == q1)]  

def unchanged(i, j, t):
   V = [k for k, (p1, p2, p3) in enumerate(TR) if (i, j) in (p1, p2, p3)]  
   cond = conjunction(y[t] != k for k in V)
   return cond == (x[t][i][j] == x[t + 1][i][j])

def to0(i, j, t):
   V = [k for k, (p1, p2, p3) in enumerate(TR) if (i, j) in (p1, p2)]
   cond = disjunction(y[t] == k for k in V)
   return cond == both(x[t][i][j] == 1, x[t + 1][i][j] == 0)

def to1(i, j, t):
   V = [k for k, (p1, p2, p3) in enumerate(TR) if (i, j) == p3]
   cond = disjunction(y[t] == k for k in V)
   return cond == both(x[t][i][j] == 0, x[t + 1][i][j] == 1)

# x[t][i](j] is the value at row i and column j at time t
x = VarArray(size=[nMoves + 1, n, m], dom=lambda t, i, j: None if init_board[i][j] is None
                                                                      else {0, 1})

# y[t] is the move (transition) performed at time t
y = VarArray(size=nMoves, dom=range(nTransitions + 1))  # +1 for NADA

# z is the smallest conflict level
z = Var(range(nMoves))

satisfy(
   # setting the initial board
   x[0] == init_board,

   # tag(symmetry-breaking)
   [(y[i], y[i + 1]) in T2 for i in range(nMoves - 1)],
   
   # tag(redundant-constraints)
   [[(y[i], y[i + 2]) not in T3 for i in range(nMoves - 2)]],
   
   # ensuring valid value of z
   y[z] == NADA,

   # ensuring no useless use of Nada
   [
      If(
         x[t][p1] == 1, x[t][p2] == 1, x[t][p3] == 0,
         Then=y[t] != NADA
      ) for k, (p1, p2, p3) in enumerate(TR) for t in range(nMoves)
   ],

   # handling unchanged situations
   [unchanged(i, j, t) for (i, j) in points for t in range(nMoves)],

   # handling situations where a peg disappears
   [to0(i, j, t) for (i, j) in points for t in range(nMoves)],
   
   # handling situations where a peg appears
   [to1(i, j, t) for (i, j) in points for t in range(nMoves)],

   # tag(redundant-constraints)
   [
      If(
         t <= z,
         Then=Sum(x[t]) == 32 - t
      ) for t in range(nMoves)
   ]
)

minimize(
   z
)
\end{python}\end{boxpy}

This model involves 2 arrays of variables, a stand-alone variable and (global) constraints of type \gb{Sum} and \gb{Table}.

A series of 17 instances has been selected for the competition.
For generating an \x3 instance (file), you can execute for example:
\begin{command}
  python FoolSolitaire.py -data=[3,3]
\end{command}

\subsection{LitPuzzle}

\paragraph{Description.}

Five puzzle, by Martin Chlond (and mentioned by Håkan Kjellerstrand on his \href{http://www.hakank.org/common_cp_models/}{website}). 
Each of the squares in the grid, below, can be in one of two states, lit (white) or unlit (red):
\begin{verbatim}
  A B C D E   
  F G H I J 
  K L M N O 
  P Q R S T 
  U V W X Y 
\end{verbatim}

If the player clicks on a square then that square and each orthogonal neighbour will toggle between the two states.
Each mouse click constitutes one move and the objective of the puzzle is to light all 25 squares in the least number of moves.
This can be generalized to a grid $n$ by $n$.

\paragraph{Data.}

The values of $n$ used for generating the 2024 competition instances are:
\begin{quote}
10, 15, 16, 17, 18, 20, 25, 30, 40, 50 
\end{quote}

\paragraph{Model.}
The \p3 model, in a file `LitPuzzle.py', used for the competition is: 

\begin{boxpy}\begin{python}
@\imp@

n = data 

# x[i,j] is 1 if the player clicks on the square at row i and column j
x = VarArray(size=[n, n], dom={0, 1})

satisfy(
   # ensuring that all cells are lit
   Sum(x.cross(i, j)) in (1, 3, 5) for i in range(n) for j in range(n)
)
   
minimize(
   Sum(x)
)
\end{python}\end{boxpy}

This model involves an array of variables $x$, and (global) constraints of type \gb{Sum}.
A series of 10 instances has been selected for the competition.
For generating an \x3 instance (file), you can execute for example:
\begin{command}
python LitPuzzle.py -data=15
\end{command}

\subsection{Maximum Density Oscillating Life}

\paragraph{Description.}

From \cite{GJMN_generating}: ``Conway’s Game of Life was invented by John Horton Conway.
The game is played on a square grid. Each cell in the grid is in one of two states (alive or dead).
The state of the board evolves over time: for each cell, its new state is determined by its previous state and the previous state of its eight neighbours (including diagonal neighbours).
Oscillators are patterns that return to their original state after a number of steps (referred to as the period).
A period 1 oscillator is named a still life.
Here we consider the problem of finding oscillators of various periods.''

\paragraph{Data.}

Two integers are required to specify a specific instance: the order $n$ of the board and the horizon $h$ allowed for solving the problem. Values of $(n,h)$ used for generating the instances in the competition are:
\begin{quote}
(5,2), (5,3), (5,4), (5,5), (5,6), (6,2), (6,3), (6,4), (6,5), (6,6), (7,2), (7,3), (7,4), (7,5), (7,6)
\end{quote}

\paragraph{Model.}

The \p3 model, in a file `MaximumDensityOscillatingLife.py', used for the \x3 competition is:

\begin{boxpy}\begin{python}
@\imp@
from pycsp3.classes.auxiliary.enums import TypeSquareSymmetry

n, horizon = data 
symmetries = [sym.apply_on(n + 2) for sym in TypeSquareSymmetry]

# x[t][i][j] is 1 iff the cell at row i and col j is alive at time t
x = VarArray(size=[horizon, n + 2, n + 2],
              dom=lambda t, i, j: {0} if i in (0, n + 1) or j in (0, n + 1) else {0, 1})

T = [(ANY, *[0] * 8, 0)] +
    [(ANY, *[1 if k == k1 else 0 for k in range(8)], 0) for k1 in range(8)] +
    [(ANY, *[0 if k in (k1, k2, k3, k4) else 1 for k in range(8)], 0)
       for k1, k2, k3, k4 in combinations(8, 4)] +
    [(ANY, *[0 if k in (k1, k2, k3) else 1 for k in range(8)], 0)
       for k1, k2, k3 in combinations(8, 3)] +
    [(ANY, *[0 if k in (k1, k2) else 1 for k in range(8)], 0)
       for k1, k2 in combinations(8, 2)] +
    [(ANY, *[0 if k == k1 else 1 for k in range(8)], 0) for k1 in range(8)] +
    [(ANY, *[1] * 8, 0)] +
    [(ANY, *[1 if k in (k1, k2, k3) else 0 for k in range(8)], 1)
       for k1, k2, k3 in combinations(8, 3)] +
    [(0, *[1 if k in (k1 k2) else 0 for k in range(8)], 0) for k1,k2 in combinations(8, 2)] +
    [(1, *[1 if k in (k1, k2) else 0 for k in range(8)], 1) for k1, k2 in combinations(8, 2)]
    
satisfy(
   # imposing rules of the game
   [
      Table(
         scope=[x[t][i][j], x[t].around(i, j), x[t + 1][i][j]],
         supports=T
      ) for t in range(horizon) for i in range(1, n + 1) for j in range(1, n + 1)
   ],

   # forbidding identical states
   AllDifferentList(x[t][1:n + 1, 1:n + 1] for t in range(horizon)),  
   
   # tag(symmetry-breaking)
   [
      [x[0] <= x[0][symmetry] for symmetry in symmetries],  
      [x[0] <= x[i] for i in range(1, horizon)]
   ]
)

maximize(
   # maximizing the number of alive cells
   Sum(x)
)
\end{python}\end{boxpy}

This model involves a three-dimensional array of variables $x$, and (global) constraints of type \gb{AllDifferentList}, \gb{Sum} and \gb{Table}.

A series of 15 instances has been selected for the competition.
For generating an \x3 instance (file), you can execute for example:
\begin{command}
  python MaximumDensityOscillatingLife.py -data=[5,2]
\end{command}

\subsection{Pyramid}

\paragraph{Description.}

Build a pyramid such as each brick of the pyramid is the sum of the two bricks situated below it.
Minimize the root value (0 not permitted) while using different values.

\paragraph{Data.}

Two integers are required to specify a specific instance: the order $n$ of the board and the limit $k$ allowed for values.
The values of ($n$,$k$) used for generating the 2024 competition instances are:
\begin{quote}
(7,500), (8,800), (9,1300), (10,2500), (11,5000), (12,10000), (13,20000), (14,45000), (15,80000), (16,200000)
\end{quote}

\paragraph{Model.}
The \p3 model, in a file `Pyramid.py', used for the competition is: 

\begin{boxpy}\begin{python}
@\imp@

n, k = data

# x[o,p] is the value of A(o,p)
x = VarArray(size=[n, n], dom=lambda i, j: range(k + 1) if j <= i else None)

satisfy(
   # imposing a non-null root
   x[0][0] != 0,

   # imposing different values
   AllDifferent(x),
   
   # computing triangle sums
   [x[i][j] == x[i + 1][j] + x[i + 1][j + 1] for i in range(n - 1) for j in range(i + 1)]
)

minimize(
   # minimizing the root value of the triangle
   x[0][0]
)
\end{python}\end{boxpy}

This model involves an array of variables $x$, and a (global) constraint of type \gb{AllDifferent}.
A series of 10 instances has been selected for the competition.
For generating an \x3 instance (file), you can execute for example:
\begin{command}
python Pyramid.py -data=[7,500]
\end{command}

\subsection{Rubiks Cube}

\paragraph{Description.}

The 1D Rubik's Cube is a vector composed of 6 number, which can be rotated in 3 different ways in groups of four:

\begin{center}
\scalebox{0.8}{
\begin{tikzpicture}[scale=0.5]
\draw [fill=gray!30] (0,0) rectangle (4,1);
\draw [very thin, gray] (0,0) grid (6,1);
\node[]() at (0.5,0.5){1};
\node[]() at (1.5,0.5){2};
\node[]() at (2.5,0.5){3};
\node[]() at (3.5,0.5){4};
\node[]() at (4.5,0.5){5};
\node[]() at (5.5,0.5){6};

\draw[->,>=stealth](7,0.5) --(9,0.5)node[above,midway]{(1)};

\def\x{10}
\draw [fill=gray!30] (0+\x,0) rectangle (4+\x,1);
\draw [very thin, gray] (0+\x,0) grid (6+\x,1);
\node[]() at (0.5+\x,0.5){4};
\node[]() at (1.5+\x,0.5){3};
\node[]() at (2.5+\x,0.5){2};
\node[]() at (3.5+\x,0.5){1};
\node[]() at (4.5+\x,0.5){5};
\node[]() at (5.5+\x,0.5){6};


\def\y{-2}

\draw [fill=gray!30] (1,0+\y) rectangle (5,1+\y);
\draw [very thin, gray] (0,0+\y) grid (6,1+\y);
\node[]() at (0.5,0.5+\y){1};
\node[]() at (1.5,0.5+\y){2};
\node[]() at (2.5,0.5+\y){3};
\node[]() at (3.5,0.5+\y){4};
\node[]() at (4.5,0.5+\y){5};
\node[]() at (5.5,0.5+\y){6};

\draw[->,>=stealth](7,0.5+\y) --(9,0.5+\y)node[above,midway]{(2)};

\draw [fill=gray!30] (1+\x,0+\y) rectangle (5+\x,1+\y);
\draw [very thin, gray] (0+\x,0+\y) grid (6+\x,1+\y);
\node[]() at (0.5+\x,0.5+\y){1};
\node[]() at (1.5+\x,0.5+\y){5};
\node[]() at (2.5+\x,0.5+\y){4};
\node[]() at (3.5+\x,0.5+\y){3};
\node[]() at (4.5+\x,0.5+\y){2};
\node[]() at (5.5+\x,0.5+\y){6};


\def\yy{-4}

\draw [fill=gray!30] (2,0+\yy) rectangle (6,1+\yy);
\draw [very thin, gray] (0,0+\yy) grid (6,1+\yy);
\node[]() at (0.5,0.5+\yy){1};
\node[]() at (1.5,0.5+\yy){2};
\node[]() at (2.5,0.5+\yy){3};
\node[]() at (3.5,0.5+\yy){4};
\node[]() at (4.5,0.5+\yy){5};
\node[]() at (5.5,0.5+\yy){6};

\draw[->,>=stealth](7,0.5+\yy) --(9,0.5+\yy)node[above,midway]{(3)};

\draw [fill=gray!30] (2+\x,0+\yy) rectangle (6+\x,1+\yy);
\draw [very thin, gray] (0+\x,0+\yy) grid (6+\x,1+\yy);
\node[]() at (0.5+\x,0.5+\yy){1};
\node[]() at (1.5+\x,0.5+\yy){2};
\node[]() at (2.5+\x,0.5+\yy){6};
\node[]() at (3.5+\x,0.5+\yy){5};
\node[]() at (4.5+\x,0.5+\yy){4};
\node[]() at (5.5+\x,0.5+\yy){3};
\end{tikzpicture}
}
\end{center}

The problem associated with the 1D Rubik's Cube can be defined in general terms: given a scrambled vector $V$ of size $n$, the objective is to return the shortest sequence of rotations (of length $g$) so as to restore the original ordered vector.
Above, we have $n=6$ and $g=4$, and the possible rotations are 1, 2, and 3 (as well as 0 for indicating that no rotation is performed). 
The number $r$ of possible rotations is equal to $n - g + 1$. 
A CP model for this problem has been proposed by H. Kjellerstrand (e.g., in Picat \cite{DBLP:conf/padl/ZhouK16}), following a post by O. Kobchenko, and the program  proposed by A. Nikitin on his MSX-BASIC page ({\footnotesize nsg.upor.net/msx/basic/line.htm}).

\paragraph{Data.}

Four integers are required to specify a specific random instance: the size $n$ of the vector, the size of the rotation slice $r$, the horizon $h$ and the seed $s$. Values of $(n,r,h,s)$ used for generating the instances in the competition are:
\begin{quote}
(10,4,20,0), (10,4,20,1), (10,4,20,2), (10,4,20,3), (11,4,20,0), (11,4,20,1), (11,4,20,2), (11,4,20,3), (12,4,20,0), (12,4,20,1), (12,4,20,2), (12,4,20,3), (13,4,20,1), (13,4,20,2)
\end{quote}

\paragraph{Model.}
The \p3 model, in a file `RubiksCube.py', used for the competition is: 

\begin{boxpy}\begin{python}
@\imp@
from random import Random

n, r, nSteps, seed = data  # seed to shuffle an initial series 
init_board = Random(seed).sample([i for i in range(1, n + 1)], n)

# final_board = list(range(1, n + 1))
nRotations = n - r + 1 + 1  # +1 for including 0 (no rotation)

# lists storing the positions to be swapped according to the operations
swaps = [[i if i not in range(op, op + r) else list(range(op + r - 1, op - 1, -1))[i - op] for i in range(n)] for op in range(nRotations - 1)]

# x[t][i] is the value of the ith element of the vector at time t
x = VarArray(size=[nSteps + 1, n], dom=range(1, n + 1))

# y[t] is the rotation chosen at time t (0 for none)
y = VarArray(size=nSteps, dom=range(nRotations))

# z is the actual number of performed rotations
z = Var(dom=range(nSteps + 1))

satisfy(
   # setting the initial board
   x[0] == init_board,

   # setting the final board
   x[-1] == range(1, n + 1),
   
   # computing when it is finished
   y[z] == 0,

   # tag(redundant-constraints)
   [AllDifferent(x[t]) for t in range(1, nSteps)]

   # ensuring valid sequences of operations
   [
     If(
        y[t] == 0,
        Then=y[t + 1] == 0,  # no more operation once finished
        Else=y[t] != y[t + 1]  # do not cancel the last operation
     ) for t in range(nSteps - 1)
   ],

   # ensuring valid rotations
   [
     (y[t] == op) == conjunction(
        x[t + 1][i] == x[t][j] for i in range(n) if (j := swaps[op - 1][i],)
     ) for t in range(nSteps - 1) for op in range(1, nRotations)
   ],

   # no more changes when finished
   [(y[t] == 0) == (x[t] == x[t + 1]) for t in range(nSteps)]
)

minimize(
   # minimizing the number of steps
   z
)
\end{python}\end{boxpy}

This model involves two arrays of variables, a stand-alone variable and (global) constraints of type \gb{AllDifferent}.
A series of 14 instances has been selected for the competition.
For generating an \x3 instance (file), you can execute for example:
\begin{command}
python RubiksCube.py -data=[10,4,20,0]
\end{command}

\subsection{SameQueensKnights}

\paragraph{Description.}

From \href{http://archive.vector.org.uk/art10003900}{archive.vector.org}: ``In 1850, Carl Friedrich Gauss and Franz Nauck showed that it is possible to place eight queens on a chessboard such that no queen attacks any other queen.
Now consider a variant of this problem: you must place a maximal number of knights and queens on a board of size $n \times n$ such that no piece attacks any other piece.''

\paragraph{Data.}

The values of $n$ used for generating the 2024 competition instances are:
\begin{quote}
5, 10, 15, 20, 25, 30, 35, 40, 45, 50, 55, 60, 65, 70, 75

\end{quote}

\paragraph{Model.}
The \p3 model, in a file `SameQueensKnights.py', used for the competition is: 

\begin{boxpy}\begin{python}
@\imp@

n = data

EMPTY, QUEEN, KNIGHT = range(3)

Cells = [(i, j) for i in range(n) for j in range(n)]

def queen_attack(i, j):
   return [(a, b) for a, b in Cells if (a, b) != (i, j)
              and (a == i or b == j or abs(i - a) == abs(j - b))]

def knight_attack(i, j):
   return [(a, b) for a, b in Cells if a != i
              and b != j and abs(i - a) + abs(j - b) == 3]

# x[i][j] indicates what is present in the cell with coordinates (i,j)
x = VarArray(size=[n, n], dom={EMPTY, QUEEN, KNIGHT})

# q is the number of queens
q = Var(dom=range(n + 1))

# k is the number of knights
k = Var(dom=range(n + 1))

satisfy(
   # computing the number of queens
   q == Sum(x[i][j] == QUEEN for i, j in Cells),

   # computing the number of knights
   k == Sum(x[i][j] == KNIGHT for i, j in Cells),

   # ensuring that no two pieces (queens or knights) attack each other
   [
     Match(
        x[i][j],
        Cases={
           QUEEN: Sum(x[queen_attack(i, j)]) == 0,
           KNIGHT: Sum(x[knight_attack(i, j)]) == 0
        }
     ) for i, j in Cells
   ]
)
        
maximize(
   q + k
)
\end{python}\end{boxpy}

This model involves an array of variables $x$, two stand-alone variables, and (global) constraints of type \gb{Sum}.
A series of 15 instances has been selected for the competition.
For generating an \x3 instance (file), you can execute for example:
\begin{command}
python SameQueensKnights.py -data=10
\end{command}

\subsection{Still Life}

This is Problem \href{https://www.csplib.org/Problems/prob032}{032} at CSPLib.

\paragraph{Description.}

From CSPLib (by Barbara Smith): ``This problem arises from the Game of Life, invented by John Horton Conway in the 1960s and popularized by Martin Gardner in his Scientific American columns.
Life is played on a squared board, considered to extend to infinity in all directions. Each square of the board is a cell,
which at any time during the game is either alive or dead. A cell has eight neighbours.
The configuration of live and dead cells at time t leads to a new configuration at time t+1 according to the rules of the game:
\begin{itemize}
\item if a cell has exactly three living neighbours at time t, it is alive at time t+1
\item if a cell has exactly two living neighbours at time t it is in the same state at time t+1 as it was at time t
\item otherwise, the cell is dead at time t+1
\end{itemize}
A stable pattern, or still-life, is not changed by these rules. Hence, every cell that has exactly three live neighbours is alive, and every cell that has fewer than two or more than three live neighbours is dead.''

Ther exists a variant `wastage'.

\paragraph{Data.}

Two integers are required to specify a specific instance: the number of rows $n$ and the number of columns $m$ of the board. Values of $(n,m)$ used for generating the instances in the competition are:
\begin{quote}
(5,5), (8,8), (8,10), (9,9), (9,12), (10,10), (10,14),(12,12), (12,18)
\end{quote}
For the variant `wastage', values of $(n,m)$ are :
\begin{quote}
(5,5), (10,10), (15,15), (20,20), (30,30), (40,40), (50,50)
\end{quote}

\paragraph{Model.}

The \p3 model, in a file `StillLife.py', used for the \x3 competition is:

\begin{boxpy}\begin{python}
@\imp@

n, m = data 

if not variant():
   T = {(v, 0) for v in range(9) if v != 3} | {(2, 1), (3, 1)}

   # x[i][j] is 1 iff the cell at row i and col j is alive
   x = VarArray(size=[n, m], dom={0, 1})

   # a[i][j] is the number of alive neighbours
   a = VarArray(size=[n, m], dom=range(9))

   satisfy(
      # computing the numbers of alive neighbours
      [Sum(x.around((i, j)) == a[i][j] for i in range(n) for j in range(m)],

      # imposing rules of the game
      [(a[i][j], x[i][j]) in T for i in range(n) for j in range(m)],

      # imposing rules for ensuring valid dead cells around the board
      [
         [x[0][i:i + 3] != (1, 1, 1) for i in range(m - 2)],
         [x[-1][i: i + 3] != (1, 1, 1) for i in range(m - 2)],
         [x[i:i + 3, 0] != (1, 1, 1) for i in range(n - 2)],
         [x[i:i + 3, - 1] != (1, 1, 1) for i in range(n - 2)]
      ],

      # tag(symmetry-breaking)
      (
          x[0][0] >= x[n - 1][n - 1],
          x[0][n - 1] >= x[n - 1][0]
      ) if n == m else None
   )

   maximize(
      # maximizing the number of alive cells
      Sum(x)
   )

elif variant("wastage"):
      
   assert n == m

   def cond_tuple(t0, t1, t2, t3, t4, t5, t6, t7, t8, wa):
      s3 = t1 + t3 + t5 + t7
      s1 = t0 + t2 + t6 + t8 + s3
      s2 = t0 * t2 + t2 * t8 + t8 * t6 + t6 * t0 + s3
      return (t4 != 1 or (2 <= s1 <= 3 and (s2 > 0 or wa >= 2) and (s2 > 1 or wa >= 1))) and \
          (t4 != 0 or (s1 != 3 and (0 < s3 < 4 or wa >= 2)) and (s3 > 1 or wa >= 1))

   T = {(*t, i) for t in product(range(2), repeat=9) for i in range(3) if cond_tuple(*t, i)}

   # x[i][j] is 1 iff the cell at row i and col j is alive (note that there is a border)
   x = VarArray(size=[n + 2, n + 2],
                  dom=lambda i, j: {0} if i in {0, n + 1} or j in {0, n + 1} else {0, 1})

   # w[i][j] is the wastage for the cell at row i and col j
   w = VarArray(size=[n + 2, n + 2], dom={0, 1, 2})

   # ws[i] is the wastage sum for cells at row i
   ws = VarArray(size=n + 2, dom=range(2 * (n + 2) * (n + 2) + 1))

   satisfy(
      # ensuring that cells at the border remain dead
      [
         [x[1][j:j + 3] != (1, 1, 1) for j in range(n)],
         [x[n][j:j + 3] != (1, 1, 1) for j in range(n)],
         [x[i:i + 3, 1] != (1, 1, 1) for i in range(n)],
         [x[i:i + 3, n] != (1, 1, 1) for i in range(n)]
      ],

      # still life + wastage constraints
      [(x[i-1:i+2, j-1:j+2], w[i][j]) in T for i in range(1, n + 1) for j in range(1, n + 1)],

      # managing wastage on the border
      [
         [(w[0][j] + x[1][j] == 1, w[n + 1][j] + x[n][j] == 1) for j in range(1, n + 1)],
         [(w[i][0] + x[i][1] == 1, w[i][n + 1] + x[i][n] == 1) for i in range(1, n + 1)]
      ],

      # summing wastage
      [Sum(w[0] if i == 0 else [ws[i - 1], w[i]]) == ws[i] for i in range(n + 2)],

      # tag(redundant)
      [ws[n + 1] - ws[i] >= 2 * ((n - i) // 3) + n // 3 for i in range(n + 1)]
   )

   maximize(
      # maximizing the number of alive cells
      (2 * n * n + 4 * n - ws[-1]) // 4
   )
\end{python}\end{boxpy}

This model involves several arrays of variables, and (global) constraints of type \gb{Sum} and \gb{Table}.

A series of $9+7$ instances has been selected for the competition.
For generating an \x3 instance (file), you can execute for example:
\begin{command}
  python StillLife.py -data=[5,5]
  python StillLife.py -variant=wastage -data=[5,5]
\end{command}

\subsection{Test Scheduling}

This is Problem \href{https://www.csplib.org/Problems/prob073}{073} on CSPLib.

\paragraph{Description.}

The problem was presented as the Industrial Modelling Challenge at the confernce CP'2015.

From CSPLib (by Morten Mossige):
``The problem arises in the context of a testing facility.
A number of tests have to be performed in minimal time.
Each test has a given duration and needs to run on one machine. While the test is running on a machine, no other test can use that machine.
Some tests can only be assigned to a subset of the machines, for others you can use any available machine.
For some tests, additional, possibly more than one, global resources are needed.
While those resources are used for a test, no other test can use the resource.
The objective is to finish the set of all tests as quickly as possible, i.e. all start times should be non-negative, and makespan should be minimized.
The makespan is the difference between the start of the earliest test, and the end of the latest finishing test.
The objective of the original industrial problem is to minimize the time required to find a schedule plus the time required to run that schedule, i.e. to minimize the time between the release of the dataset and the conclusion of all tests required.
As this objective depends on the speed of the machine(s) on which the schedule is generated, it is hard to compare results in an objective fashion.''

\paragraph{Data.}

As an illustration of data specifying an instance of this problem, we have:
\begin{json}
{
  "nMachines": 10,
  "resourceCapacities": [1, 1, 1, 1, 1, 1, 1, 1, 1, 1],
  "tests": [
    {
      "duration": 328,
      "machines": [4, 3],
      "resources": []
    },
    {
      "duration": 339,
      "machines": [0, 1, 2, 3, 4, 5, 6, 7, 8, 9],
      "resources": []
    },
    ...
    {
      "duration": 506,
      "machines": [0, 1, 2, 3, 4, 5, 6, 7, 8, 9],
      "resources": [5, 1, 0]
    }
  ]
}
\end{json}

\paragraph{Model.}

The \p3 model, in a file `TestScheduling.py' used for the \x3 competition is: 

\begin{boxpy}\begin{python}
@\imp@

nMachines, nResources, tests = data
durations, machines, resources = zip(*tests)  # information split over the tests
nTests = len(tests)

horizon = sum(durations) + 1  

tests_by_single_machines = [t for t in [[i for i in range(nTests) if len(machines[i]) == 1 and m in machines[i]] for m in range(nMachines)] if len(t) > 1]
tests_by_resources = [t for t in [[i for i in range(nTests) if r in resources[i]] for r in range(nResources)] if len(t) > 1]

def conflicting_tests():
   def possibly_conflicting(i, j):
      return len(machines[i]) == 0 or len(machines[j]) == 0
              or len(set(machines[i] + machines[j])) != len(machines[i]) + len(machines[j])

   pairs = [(i, j) for i, j in combinations(range(nTests), 2) if possibly_conflicting(i, j)]
   for t in tests_by_single_machines + tests_by_resources:
      for i, j in combinations(t, 2):
         if (i, j) in pairs:
            pairs.remove((i, j))  # because will be considered in another posted constraint
   return pairs

# s[i] is the starting time of the ith test
s = VarArray(size=nTests, dom=range(horizon))

# m[i] is the machine used for the ith test
m = VarArray(size=nTests,
               dom=lambda i: range(nMachines) if len(machines[i]) == 0 else machines[i])

satisfy(
   # no overlapping on machines
   [
      If(
         m[i] == m[j],
         Then=either(s[i] + durations[i] <= s[j], s[j] + durations[j] <= s[i])
      ) for i, j in conflicting_tests()
   ],

   # no overlapping on single pre-assigned machines
   [
      NoOverlap(
         tasks=[Task(origin=s[i], length=durations[i]) for i in t]
      ) for t in tests_by_single_machines
   ],

   # no overlapping on resources
   [
      NoOverlap(
         tasks=[Task(origin=s[i], length=durations[i]) for i in t]
      ) for t in tests_by_resources
   ],

   # no more than the available machines available at any time  tag(redundant)
   Cumulative(
      origins=s,
      lengths=durations,
      heights=[1] * nTests
   ) <= nMachines
)

minimize(
   # minimizing the makespan
   Maximum(s[i] + durations[i] for i in range(nTests))
)
\end{python}\end{boxpy}

This model involves two arrays of variables, and (global) constraints of type \gb{Cumulative}, \gb{Maximum}, and \gb{NoOverlap}.

A series of 21 instances has been selected for the competition.
For generating an \x3 instance (file), you can execute for example:
\begin{command}
python TestScheduling.py -data=<datafile.json>
python TestScheduling.py -data=<datafile.pl> -parser=TestScheduling_Parser.py
\end{command}

\subsection{Traveling Tournament}

This problem is related to Problem \href{https://www.csplib.org/Problems/prob068/}{068} on CSPLib.
Many relevant information can be found in \href{https://www.researchgate.net/publication/220270875_The_Traveling_Tournament_Problem_Description_and_Benchmarks}{this paper}.
See also \cite{ENT_solving}.

\paragraph{Description.}

\begin{quote}
``The Traveling Tournament Problem (TTP) is defined as follows.
  A double round robin tournament is played by an even number of teams.
  Each team has its own venue at its home city.
  All teams are initially at their home cities, to where they return after their last away game.
  The distance from the home city of a team to that of another team is known beforehand.
  Whenever a team plays two consecutive away games, it travels directly from the venue of the first opponent to that of the second.
  The problem calls for a schedule such that no team plays more than (two or) three consecutive home games or more than (two or) three consecutive away games,
  there are no consecutive games involving the same pair of teams, and the total distance traveled by the teams during the tournament is minimized.''
\end{quote}

\paragraph{Data.}

As an illustration of data specifying an instance of this problem, we have:
\begin{json}
{
  "distances": [
    [0, 10, 15, 34],
    [10, 0, 22, 32],
    [15, 22, 0, 47],
    [34, 32, 47, 0]
  ]
}
\end{json}

\paragraph{Model.}

The \p3 model, in a file `TravelingTournament.py' used for the \x3 competition is: 

\begin{boxpy}\begin{python}
@\imp@

istances = data
nTeams, nRounds = len(distances), len(distances) * 2 - 2
assert nTeams % 2 == 0, "An even number of teams is expected"
nConsecutiveGames = 2 

def T3(i):  # this is a table for the first or last game of the ith team
   return {(1, ANY, 0)} | {(0, j, distances[i][j]) for j in range(nTeams) if j != i}

def T5(i):  # this is a table for a game that is not the first or last one of the ith team
   return ({(1, 1, ANY, ANY, 0)} |
           {(0, 1, j, ANY, distances[i][j]) for j in range(nTeams) if j != i} |
           {(1, 0, ANY, j, distances[i][j]) for j in range(nTeams) if j != i} |
           {(0, 0, j1, j2, distances[j1][j2]) for j1 in range(nTeams) for j2 in range(nTeams) if different_values(i, j1, j2)})

def A():
   qi, q01, q02, q03, q11, q12, q13 = states = "q", "q01", "q02", "q03", "q11", "q12", "q13"
   tr = [(qi, 0, q01), (qi, 1, q11), (q01, 0, q02), (q01, 1, q11), (q11, 0, q01),
           (q11, 1, q12), (q02, 1, q11), (q12, 0, q01)]
   return Automaton(start=qi, final=states[1:], transitions=tr)

# o[i][k] is the opponent (team) of the ith team  at the kth round
o = VarArray(size=[nTeams, nRounds], dom=range(nTeams))

# h[i][k] is 1 iff the ith team plays at home at the kth round
h = VarArray(size=[nTeams, nRounds], dom={0, 1})

# t[i][k] is the traveled distance by the ith team at the kth round. An additional round is considered for returning home.
t = VarArray(size=[nTeams, nRounds + 1], dom=distances)

satisfy(
   # each team must play exactly two times against each other team
   [
      Cardinality(
         within=o[i],
         occurrences={j: 2 for j in range(nTeams) if j != i}
      ) for i in range(nTeams)
   ],

   # if team i plays against j at round k, then team j plays against i at round k
   [o[o[i][k]][k] == i for i in range(nTeams) for k in range(nRounds)],

   # channeling the arrays o and h
   [h[o[i][k]][k] != h[i][k] for i in range(nTeams) for k in range(nRounds)],

   # playing against the same team must be done once at home and once away
   [
      If(
         o[i][k1] == o[i][k2],
         Then=h[i][k1] != h[i][k2]
      ) for i in range(nTeams) for k1, k2 in combinations(nRounds, 2)
   ],

   # at each round, opponents are all different  tag(redundant)
   [AllDifferent(o[:, k]) for k in range(nRounds)],
   
   # tag(symmetry-breaking)
   o[0][0] < o[0][-1],
   
   # at most 'nConsecutiveGames' consecutive games at home, or consecutive games away
   [h[i] in A() for i in range(nTeams)],
   
   # handling traveling for the first game
   [(h[i][0], o[i][0], t[i][0]) in T3(i) for i in range(nTeams)],
   
   # handling traveling for the last game
   [(h[i][-1], o[i][-1], t[i][-1]) in T3(i) for i in range(nTeams)],
   
   # handling traveling for two successive games
   [(h[i][k], h[i][k + 1], o[i][k], o[i][k + 1], t[i][k + 1]) in T5(i)
       for i in range(nTeams) for k in range(nRounds - 1)]
)

minimize(
   # minimizing summed up traveled distance
   Sum(t)
)
\end{python}\end{boxpy}

This model involves three arrays of variables, and (global) constraints of type \gb{AllDifferent}, \gb{Cardinality}, \gb{Element}, \gb{Regular}, \gb{Sum} and \gb{Table}.

A series of 14 instances has been selected for the competition.
For generating an \x3 instance (file), you can execute for example:
\begin{command}
python TravelingTournament.py -data=<datafile.json>
python TravelingTournament.py -data=<datafile.pl> -parser=TravelingTournament_Parser.py
\end{command}

\subsection{Vehicle Routing Problem with Location Congestion}

\paragraph{Description.}

From \cite{LV_branch}: ``A Vehicle Routing Problem (VRP) is a combinatorial optimization problem that aims to construct routes for a fleet of vehicles that service customer requests while minimizing some cost function.
The family of VRPs is extensive and includes variants that specify additional side constraints, such as time window constraints that restrict the time at which service of a request can commence, and precedence constraints that require one request to be serviced before another.
The variant named the Vehicle Routing Problem with Pickup and Delivery, Time Windows, and Location Congestion (VRP-PDTWLC, or VRPLC for short), is motivated by applications in humanitarian and military logistics, where Air Force bases have limited parking spots, fuel reserve, and landing and takeoff times for airplane operations.''

\paragraph{Data.}

As an illustration of data specifying an instance of this problem, we have:
\begin{json}
{
  "T": 150,
  "V": 10,
  "Q": 15,
  "L": 5,
  "C": 1,
  "P": 10,
  "times": [
    [0, 4, ..., 18],
    [4, 0, ..., 17],
    ...
    [18, 17, 0]
  ],
  "requests": [
    {"l": 0, "a": 18, "b": 52, "s": 14, "q": 1},
    {"l": 2, "a": 24, "b": 46, "s": 13, "q": 3},
    ...
    {"l": 1, "a": 64, "b": 119, "s": 15, "q": -1}
  ]
}
\end{json}

\paragraph{Model.}

An original model was proposed by Edward Lam, and described in \cite{LV_branch}.
The \p3 model, in a file `VRP\_LC.py', used for the \x3 competition is close to (can be seen as the close translation of) the one submitted to the 2018 Minizinc challenge.

\begin{boxpy}\begin{python}
@\imp@

horizon, nVehicles, vehCapacity, nLocations, locCapacity, nPickups, times, requests = data
rl, ra, rb, rs, rq = zip(*requests)
nNodes, n = len(times), len(requests)  # n is the number of requests (pickups and deliveries)

MainNodes = range(n + nVehicles)  # request and start nodes
Depots = range(n, nNodes)  # start and end nodes
load_changes = cp_array(list(rq) + [0 for _ in Depots])

# veh[i] is the vehicle visiting the ith node
veh = VarArray(size=nNodes, dom=range(nVehicles))

# succ[i] is the node that succeeds to the ith node
succ = VarArray(size=nNodes, dom=range(nNodes))

# load[i] is the load after visiting the ith node
load = VarArray(size=nNodes, dom=range(vehCapacity + 1))

# arr[i] is the arrival time at the ith node
arr = VarArray(size=nNodes, dom=range(horizon + 1))

# ser[i] is the starting time of service at the ith node
ser = VarArray(size=nNodes, dom=range(horizon + 1))

# dep[i] is the departure time of the ith node
dep = VarArray(size=nNodes, dom=range(horizon + 1))

satisfy(
   # ensuring a circuit
   Circuit(succ),

   # giant tour representation of routes
   (
      [succ[n + nVehicles + v] == n + v + 1 for v in range(nVehicles - 1)],
      succ[-1] == n
   ),

   # tracking vehicle along route
   (
      [veh[i] == veh[succ[i]] for i in MainNodes],
      [veh[n + v] == v for v in range(nVehicles)],
      [veh[n + nVehicles + v] == v for v in range(nVehicles)]
   ),

   # ordering time-related variables
   (
      (
         arr[i] <= ser[i],
         ser[i] + rs[i] <= dep[i],
         ra[i] <= ser[i],
         ser[i] <= rb[i]
      ) for i in range(n)
   ),

   # setting time at start and end nodes
   [
      [arr[i] == ser[i] for i in Depots],
      [ser[i] == dep[i] for i in Depots]
   ],

   # accumulating time along route
   [dep[i] + times[i][succ[i]] == arr[succ[i]] for i in MainNodes],

   # accumulating load along route
   [load[i] + load_changes[succ[i]] == load[succ[i]] for i in MainNodes],
   
   # setting load at start and end nodes
   [load[i] == 0 for i in Depots],
   
   # delivery after pickup
   [dep[i] + times[i][nPickups + i] <= arr[nPickups + i] for i in range(nPickups)],
   
   # delivery on same vehicle as pickup
   [veh[i] == veh[nPickups + i] for i in range(nPickups)],
   
   # handling service resources
   [
      Cumulative(
         tasks=[Task(origin=ser[i], length=rs[i], height=1) for i in range(n) if rl[i] == p]
      ) <= locCapacity for p in range(nLocations)
   ],

   # tag(symmetry-breaking)
   (
      [
         If(
            succ[n + v] == n + nVehicles + v,
            Then=succ[n + v + 1] == n + nVehicles + v + 1
         ) for v in range(nVehicles - 1)
      ],
      veh[0] == 0
   )
)

minimize(
   # minimizing the total travel distance
   Sum(times[i][succ[i]] for i in MainNodes)
)
\end{python}\end{boxpy}

This model involves six arrays of variables and (global) constraints of type \gb{Circuit}, \gb{Cumulative}, \gb{Element} and \gb{Sum}.
A series of 10 instances has been selected for the competition.
For generating an \x3 instance (file), you can execute for example:
\begin{command}
python VRP_LC.py -data=inst.json
\end{command}
where `inst.json' is a data file in JSON format.
Or you can execute: 

\begin{command}
python VRP_LC.py -data=inst.dzn -parser=VRP_LC_ParserZ.py
\end{command}
where `inst.dzn' is a data Zinc file.

\subsection{Word Golf}

\paragraph{Description.}

From \href{https://en.wikipedia.org/wiki/Word_ladder}{Wikipedia}: ``Word ladder (also known as Doublets, word-links, change-the-word puzzles, paragrams, laddergrams, or word golf) is a word game invented by Lewis Carroll.
A word ladder puzzle begins with two words, and to solve the puzzle one must find a chain of other words to link the two, in which two adjacent words (that is, words in successive steps) differ by one letter.''

\paragraph{Data.}

In addition to the name of a dictionary file, three integers are required to specify a specific instance: the size of the words $m$, the number of allowed steps $h$ and the seed $s$. Values of $(n,h,s)$ used for generating the instances in the competition are:
\begin{quote}
(4,50,0), (4,50,1), (4,50,2), (4,50,3), (5,50,0), (5,50,1), (5,50,2), (5,50,3), (6,50,0), (6,50,1), (6,50,2), (6,50,3), (7,50,0), (7,50,1), (7,50,2), (7,50,3)
\end{quote}

\paragraph{Model.}

The \p3 model, in a file `WordGolf.py', used for the \x3 competition is:

\begin{boxpy}\begin{python}
@\imp@

m, dict_name, nSteps, seed = data

words = []
for line in open(dict_name):
   code = alphabet_positions(line.strip().lower())
   if len(code) == m:
      words.append(code)
words, nWords = cp_array(words), len(words)

random.seed(seed)
start, end = random.randint(0, nWords // 2), random.randint(nWords // 2, nWords)
print(words[start], words[end])

#  x[i][j] is the letter, number from 0 to 25, at row i and column j
x = VarArray(size=[nSteps, m], dom=range(26))

# y[i] is the word index of the ith word
y = VarArray(size=nSteps, dom=range(nWords))

# z is the number of steps
z = Var(range(nSteps))

satisfy(
   # setting the start word
   [
      x[0] == words[start],
      y[0] == start
   ],

   # setting the end word
   [
      x[-1] == words[end],
      y[-1] == end
   ],

   # setting the ith word
   [x[i] == words[y[i]] for i in range(1, nSteps - 1)]

   # ensuring a (Hamming) distance of 1 between two successive words
   [
      If(
         i < z,
         Then=Hamming(x[i], x[i + 1]) == 1,
         Else=y[i] == y[i + 1]
      ) for i in range(nSteps - 1)
   ]
   
   # setting the objective value (number of steps)
   y[z] == end
)

minimize(
   z
)
\end{python}\end{boxpy}

This model involves tw arrays of variables, a stand-alone variable and (global) constraints of type \gb{Count} (generated from function 'Hamming') and \gb{Element}.

A series of $16$ instances has been selected for the competition.
For generating an \x3 instance (file), you can execute for example:
\begin{command}
python WordGolf.py -data=[7,ogd2008,20,3] 
\end{command}

wher `ogd2008' is the name of a dictionary file.

\subsection{Wordpress}

\paragraph{Description.}

From \cite{EMZ_scalable}: ``The problem of Cloud resource provisioning for component-based applications consists in the allocation of virtual machines (VMs) offers from various Cloud Providers (CPs), to a set of applications such that the constraints induced by the interactions between components and by the components hardware/software requirements are satisfied and the performance objectives are optimized (e.g. costs are minimized).
This problem has some connection with the bin-packing problem''.

\paragraph{Data.}

As an illustration of data specifying an instance of this problem, we have:
\begin{json}
{
  "WPinstance": 7,
  "nWMs": 15,
  "requirementsPerComponent": [
    [2, 512, 1000],
    [2, 512, 2000],
    ...,
    [4, 4000, 500]
  ],
  "offers": [
    [64, 976000, 1000],
    [64, 488000, 8000],
    ...,
    [0, 0, 0]
  ],
  "prices": [ 8403, 9152, ..., 0]
}
\end{json}

\paragraph{Model.}

An original model was originally designed and implemented by Andrei Iovescu, and adapted by David Bogdan for the 2022 Minizinc Challenge.
The \p3 model, in a file `Wordpress.py', used for the \x3 competition is close to (can be seen as the close translation of) the one submitted to the 2022 Minizinc challenge.

\begin{boxpy}\begin{python}
@\imp@

bWP, nVMs, requirements, types, prices = data  
nComponents, nFeatures = 5, len(requirements[0])  # features are hardware settings 
nTypes = len(types)  # note that a dummy type has been added at last position by the parser
assert len(requirements) == 5

# WP for WordPress, SQL for MySQL, DNS for DNS_LoadBalancer,
# HTTP for HTTP_LoadBalancer, VS for Varnish Software
WP, SQL, DNS, HTTP, VS = range(nComponents)

# x[i][k] is 1 iff the kth VM is used for the ith component
x = VarArray(size=[nComponents, nVMs], dom={0, 1})

# oc[k] is 1 iff the kth VM is used/occupied (i.e., not a dummy VM)
oc = VarArray(size=nVMs, dom={0, 1})

# tp[k] is the type of the kth chosen VM
tp = VarArray(size=nVMs, dom=range(nTypes))

# pr[k] is the price of the kth chosen VM
pr = VarArray(size=nVMs, dom=range(0, 16001))


satisfy(
   # ensuring certain limits (cardinality) on the deployment of components
   [
      Sum(x[WP]) >= lbWP,
      Sum(x[SQL]) >= 2,
      Sum(x[VS]) >= 2,
      Sum(x[DNS]) <= 1
   ],

   # ensuring used VMs are considered as being occupied
   [oc[k] == (Sum(x[:, k]) > 0) for k in range(nVMs)],

   # tag(symmetry-breaking)
   Decreasing(oc),

   # ensuring that hardware requirements are met for each component
   [x[:, k] * requirements[:, h] <= types[tp[k], h]
       for k in range(nVMs) for h in range(nFeatures)],

   # computing prices of chosen VMs
   [pr[k] == prices[tp[k]] for k in range(nVMs)],

   # ensuring certain connections between deployment of components
   [
      (Sum(x[DNS]) > 0) != (Sum(x[HTTP]) > 0),
      If(
         Sum(x[DNS]) > 0,
         Then=Sum(x[WP]) <= 7 * Sum(x[DNS]),
         Else=Sum(x[WP]) <= 3 * Sum(x[HTTP])
      ),
      2 * Sum(x[WP]) <= 3 * Sum(x[SQL])
   ],

   # ensuring that conflicting components do not share the same VM
   [
      [x[VS][k] + x[i][k] <= 1 for k in range(nVMs) for i in (DNS, HTTP, SQL)],
      [x[DNS][k] + x[i][k] <= 1 for k in range(nVMs) for i in (WP, SQL, VS)],
      [x[HTTP][k] + x[i][k] <= 1 for k in range(nVMs) for i in (WP, SQL, VS)]
   ]
)

minimize(
   # minimizing overall cost
   Sum(pr)
)
\end{python}\end{boxpy}

This model involves four arrays of variables and (global) constraints of type \gb{Element} and \gb{Sum}.

A series of 13 instances has been selected for the competition.
For generating an \x3 instance (file), you can execute for example:
\begin{command}
python Wordpress.py -data=inst.json
\end{command}
where `inst.json' is a data file in JSON format.

\chapter{Solvers}

In this chapter, we introduce the solvers and teams having participated to the \x3 Competition 2024.

\begin{itemize}
\item ACE (Christophe Lecoutre)
\item BTD, miniBTD (Mohamed Sami Cherif, Djamal Habet, Philippe J\'egou, H\'el\`ene Kanso, Cyril Terrioux)
\item Choco (Charles Prud'homme)
\item CoSoCo (Gilles Audemard)
\item CPMpy {\small (T. Guns, W.Vanroose, T. Sergeys, I.Bleukx, J. Devriendt, D. Tsouros, H. Verhaeghe)} \\
  cpmpy\_mzn\_chuffed, cpmpy\_mzn\_gecode, cpmpy\_z3, cpmpy\_exact, ppmpy\_gurobi, cpmpy\_ortools
\item Exchequer (Martin Mariusz Lester)
\item Fun-sCOP (Takehide Soh, Daniel Le Berre, Hidetomo Nabeshima, Mutsunori Banbara, Naoyuki Tamura)
\item Nacre (Ga\"el Glorian)
\item Picat (Neng-Fa Zhou)
\item RBO, miniRBO (Mohamed Sami Cherif, Djamal Habet, Cyril Terrioux)
\item Sat4j-CSP-PB (extension of Sat4j by Thibault Falque and Romain Wallon)
\item toulbar2 (David Allouche et al.) 
\end{itemize}

\addcontentsline{toc}{section}{\numberline{}ACE}
\includepdf[pages=-,pagecommand={\thispagestyle{plain}}]{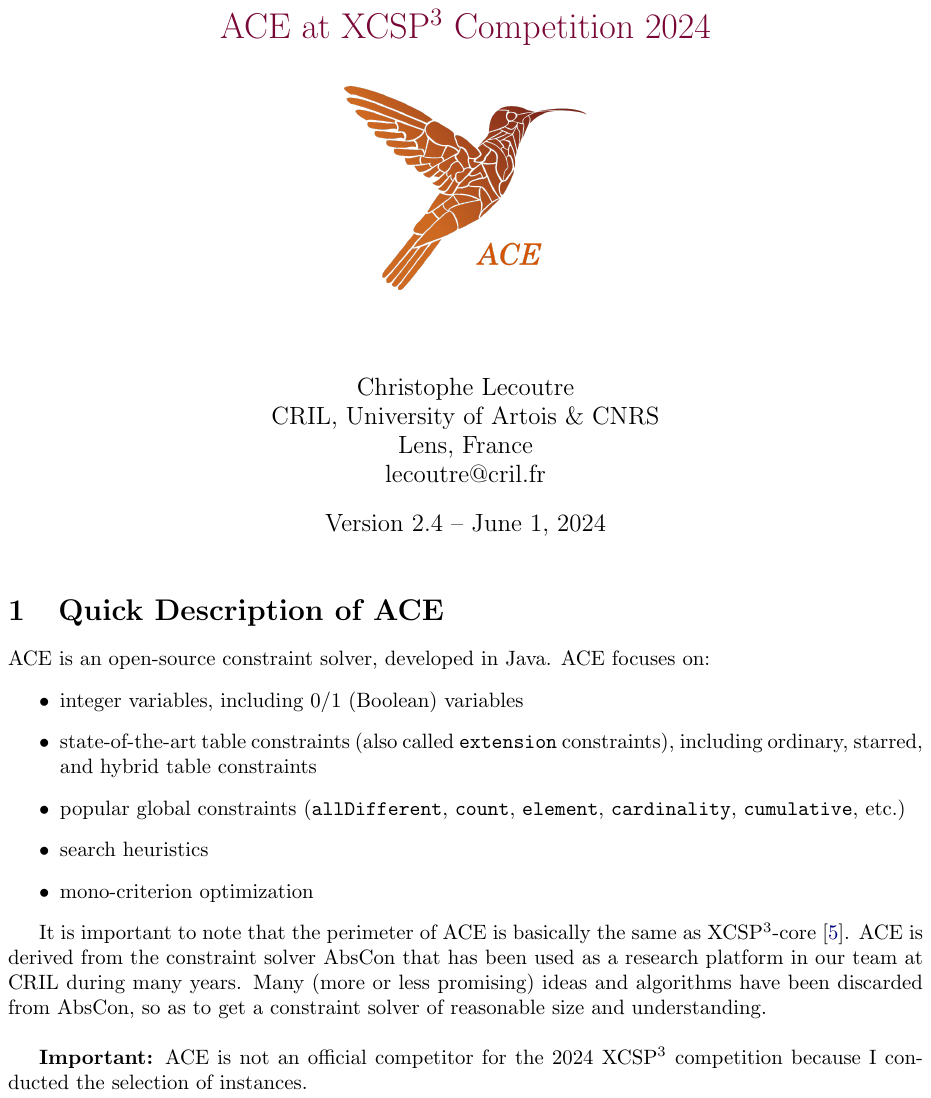}
\addcontentsline{toc}{section}{\numberline{}BTD}
\includepdf[pages=-,pagecommand={\thispagestyle{plain}}]{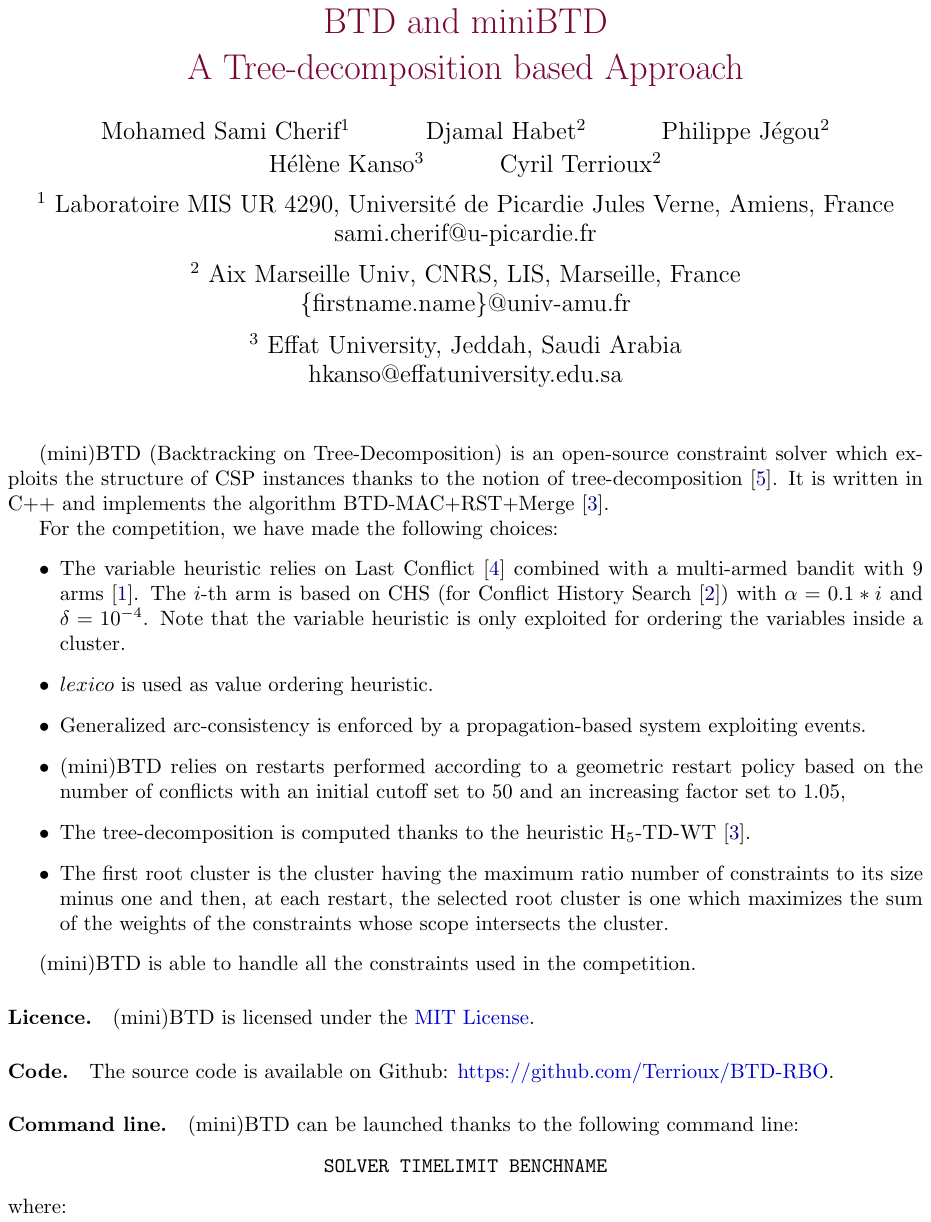}
\addcontentsline{toc}{section}{\numberline{}Choco}
\includepdf[pages=-,pagecommand={\thispagestyle{plain}}]{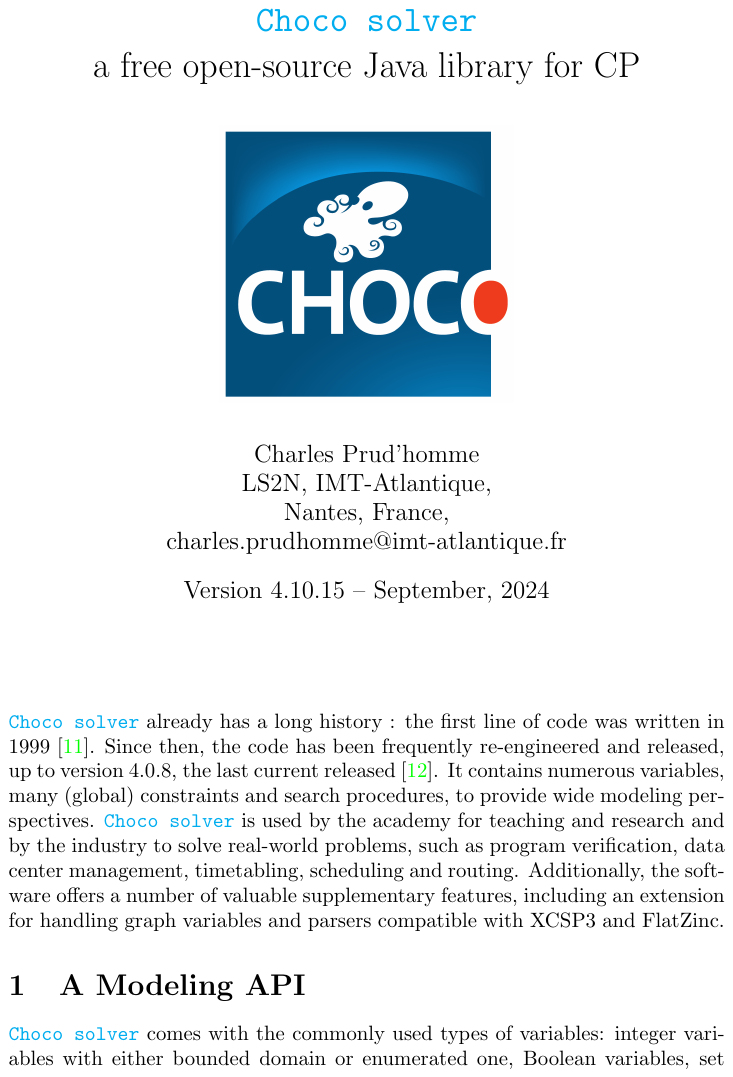}
\addcontentsline{toc}{section}{\numberline{}CoSoCo}
\includepdf[pages=-,pagecommand={\thispagestyle{plain}}]{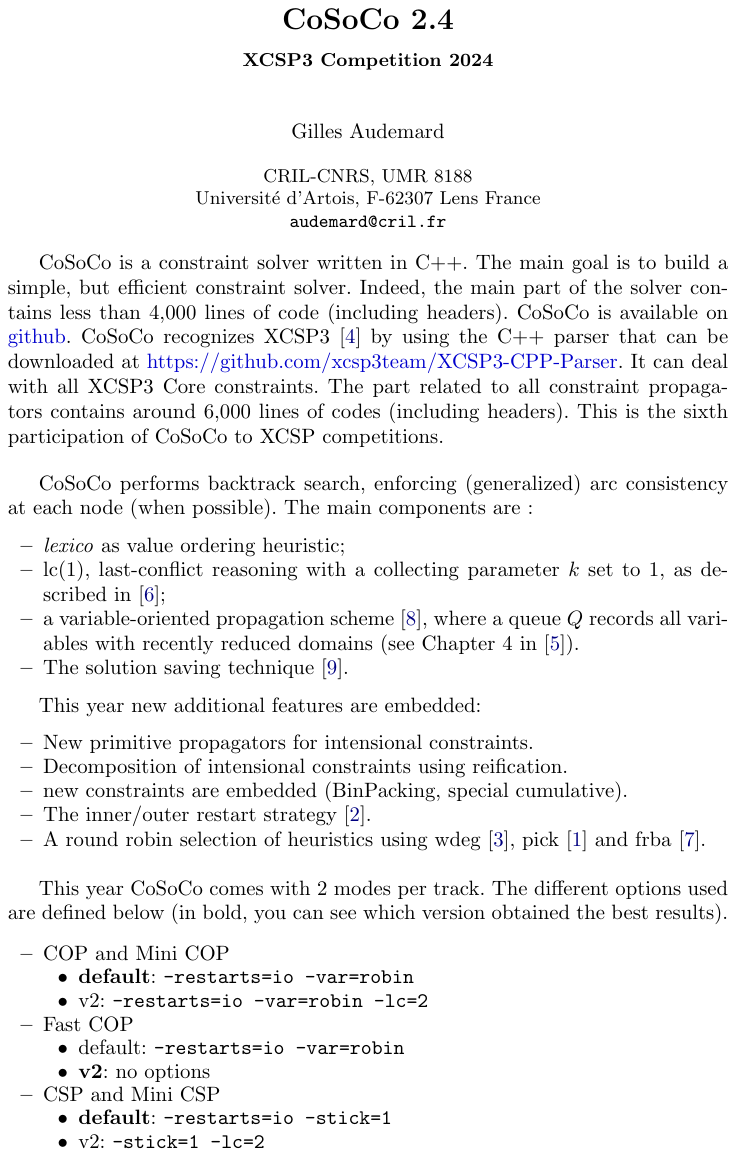}
\addcontentsline{toc}{section}{\numberline{}CPMpy}
\includepdf[pages=-,pagecommand={\thispagestyle{plain}}]{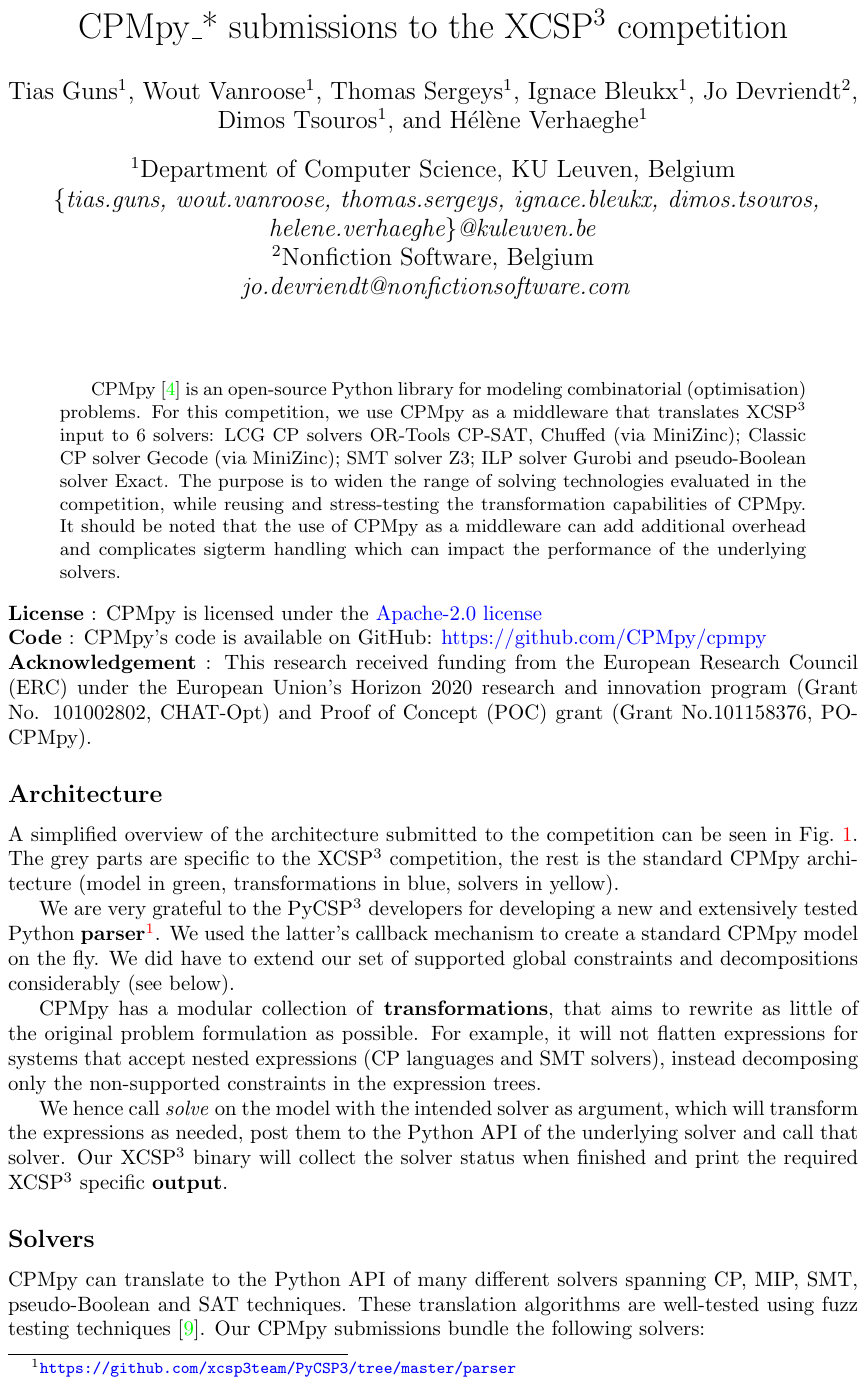}
\addcontentsline{toc}{section}{\numberline{}Exchequer}
\includepdf[pages=-,pagecommand={\thispagestyle{plain}}]{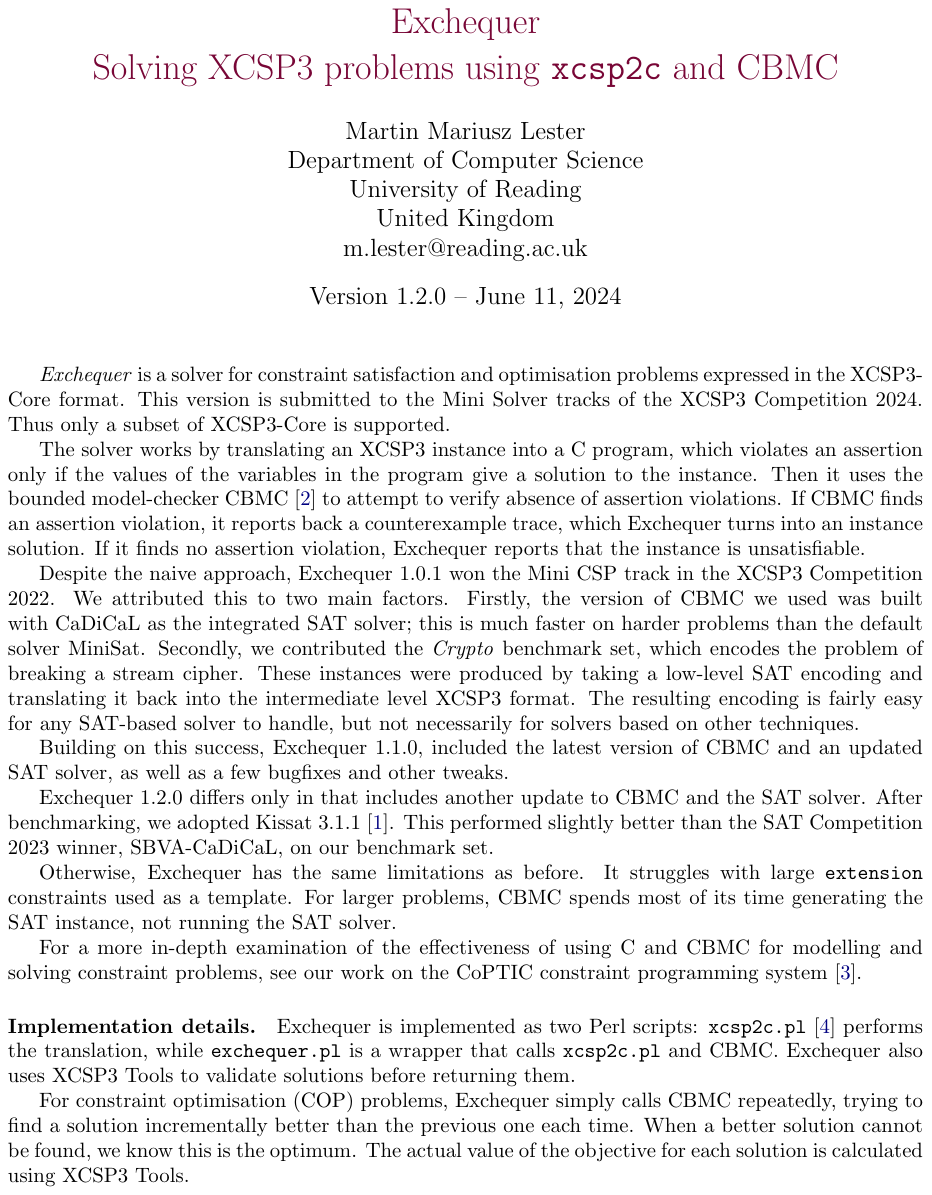}
\addcontentsline{toc}{section}{\numberline{}Fun-sCOP}
\includepdf[pages=-,pagecommand={\thispagestyle{plain}}]{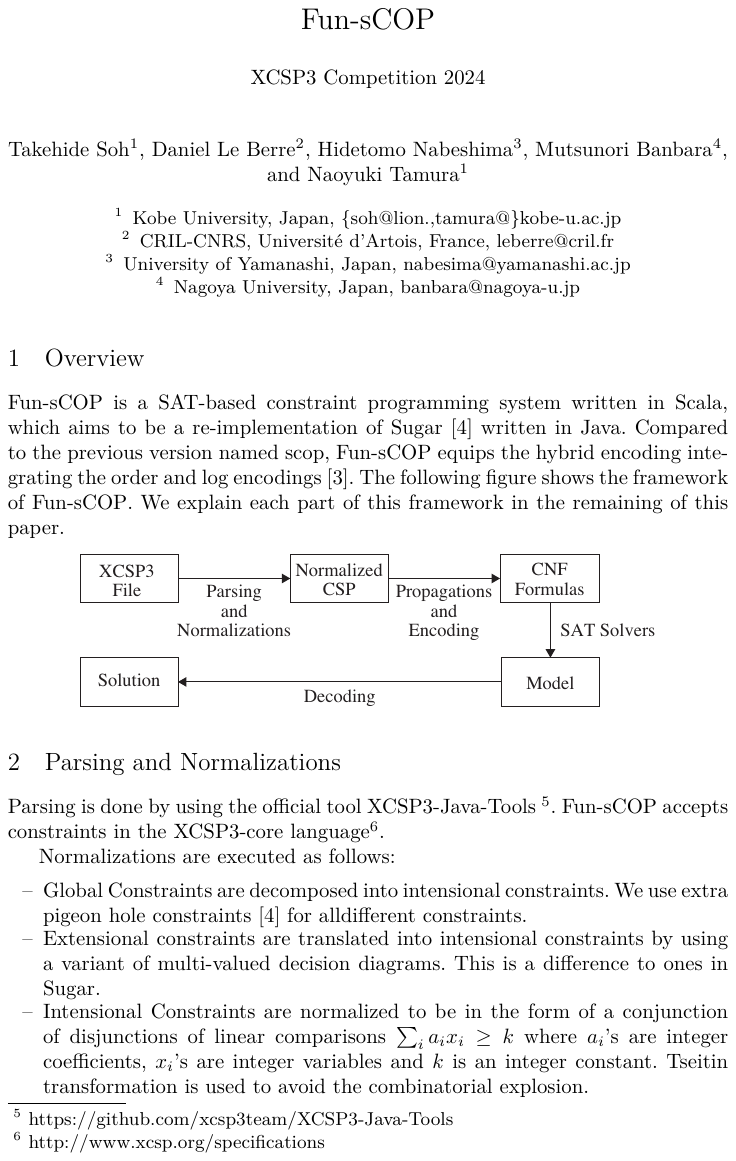}
\addcontentsline{toc}{section}{\numberline{}NACRE}
\includepdf[pages=-,pagecommand={\thispagestyle{plain}}]{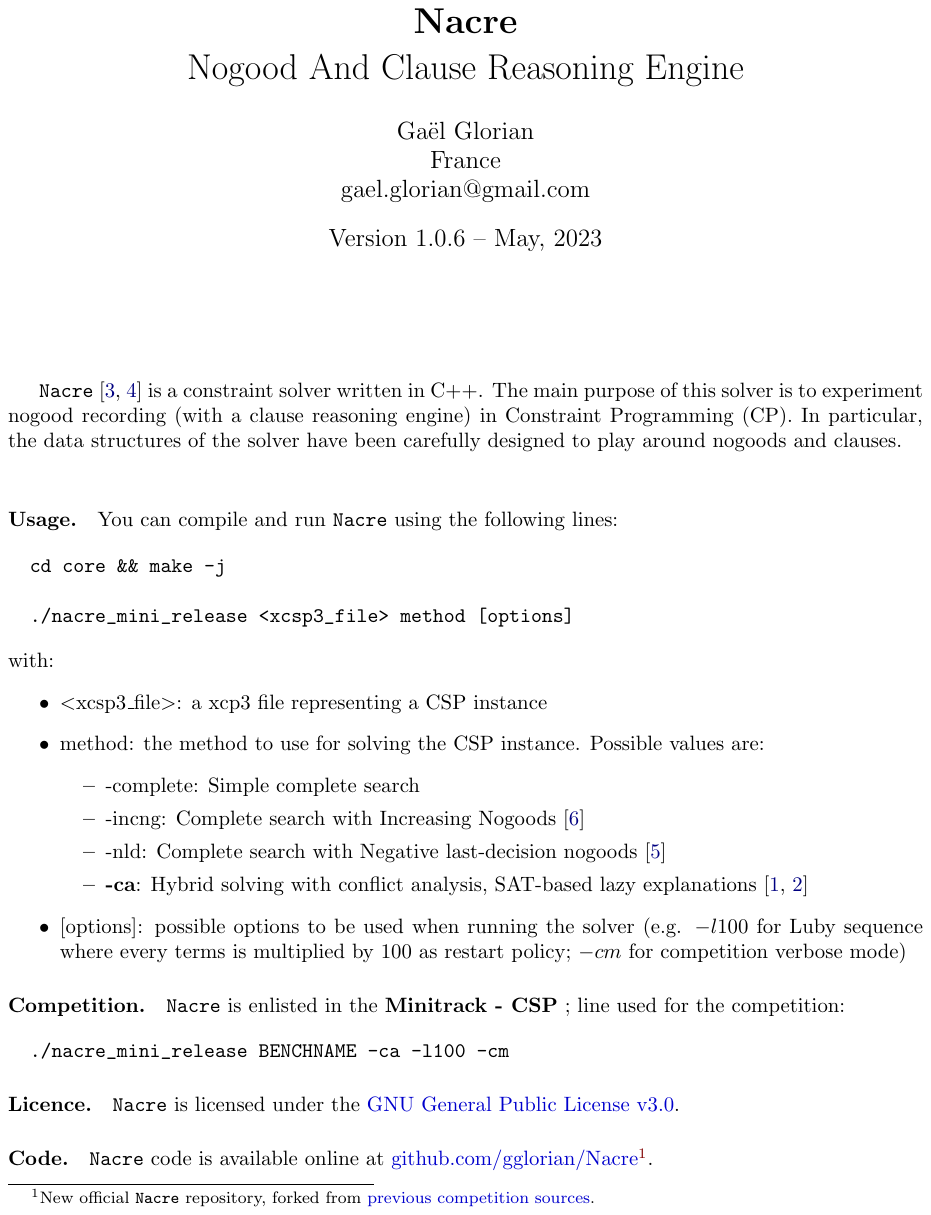}
\addcontentsline{toc}{section}{\numberline{}Picat}
\includepdf[pages=-,pagecommand={\thispagestyle{plain}}]{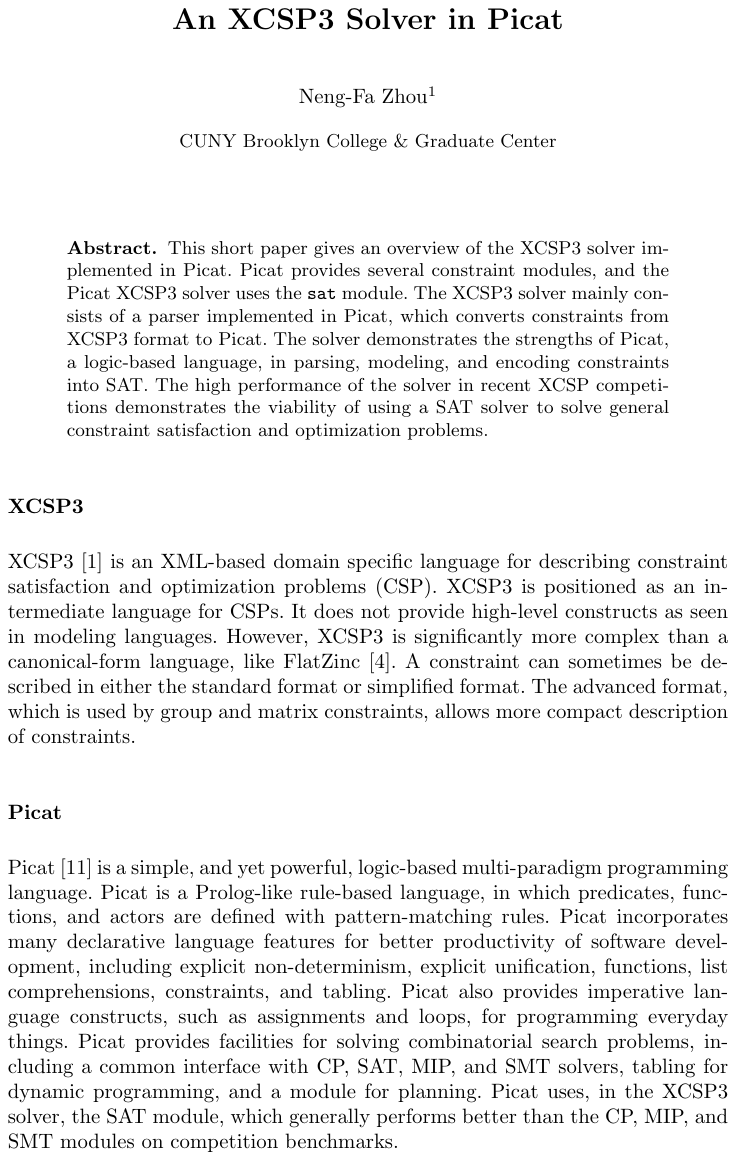}
\addcontentsline{toc}{section}{\numberline{}RBO}
\includepdf[pages=-,pagecommand={\thispagestyle{plain}}]{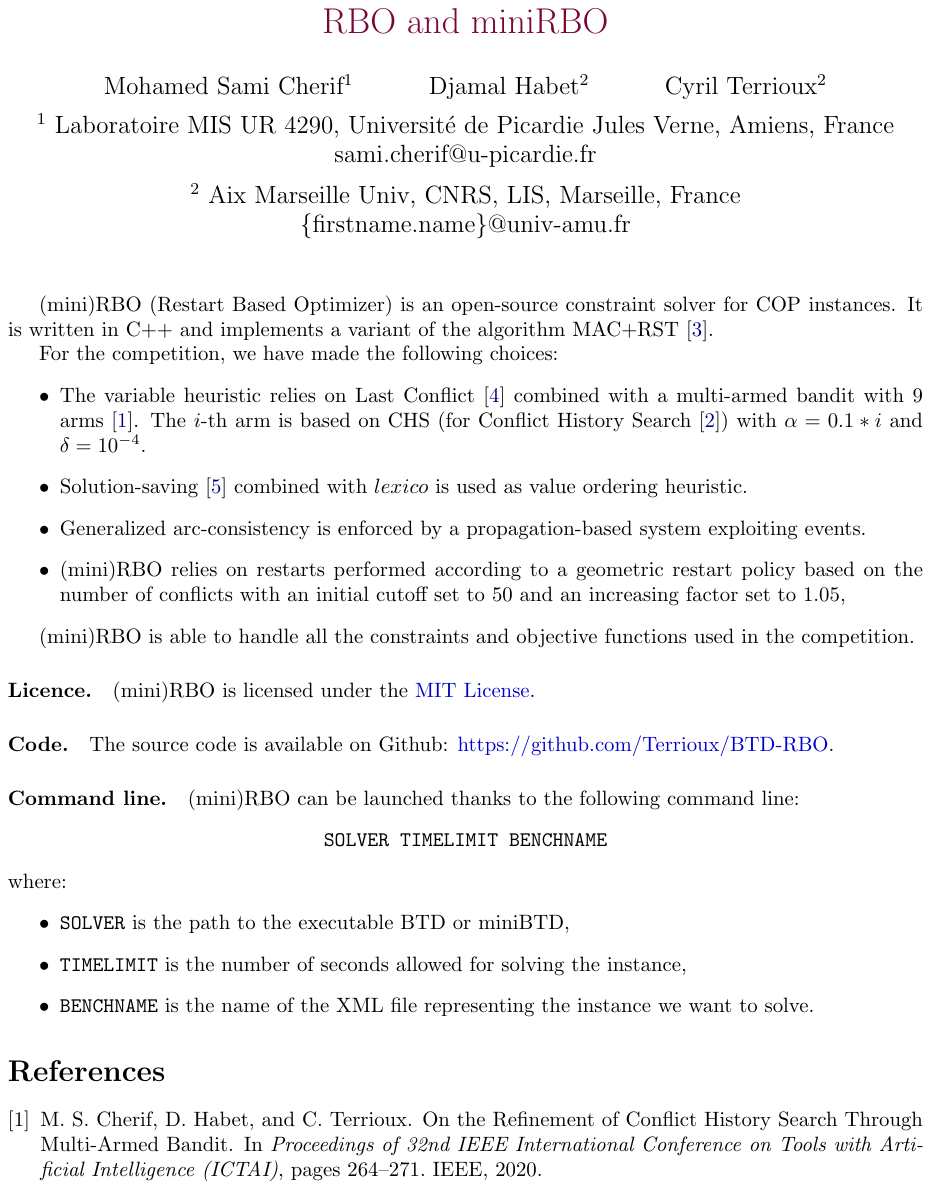}
\addcontentsline{toc}{section}{\numberline{}Sat4j-CSP-PBj}
\includepdf[pages=-,pagecommand={\thispagestyle{plain}}]{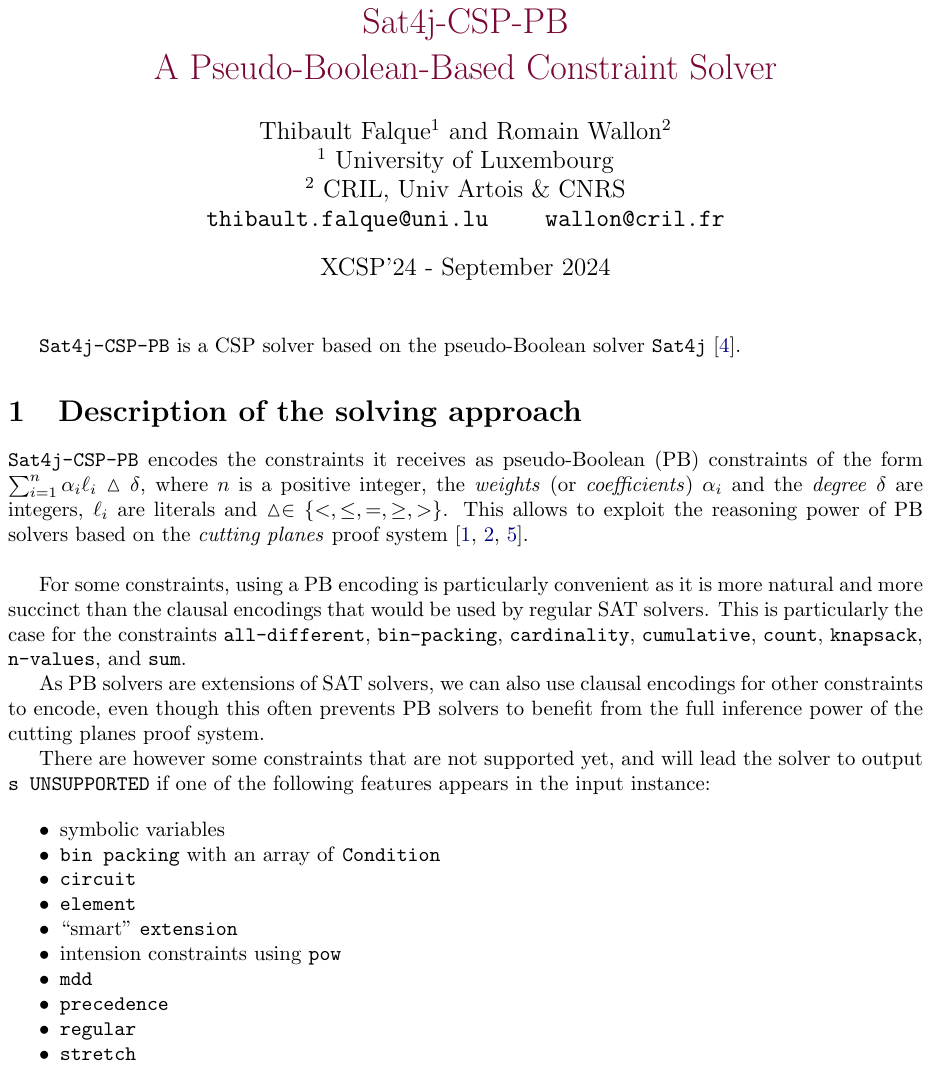}
\addcontentsline{toc}{section}{\numberline{}toulbar2}
\includepdf[pages=-,pagecommand={\thispagestyle{plain}}]{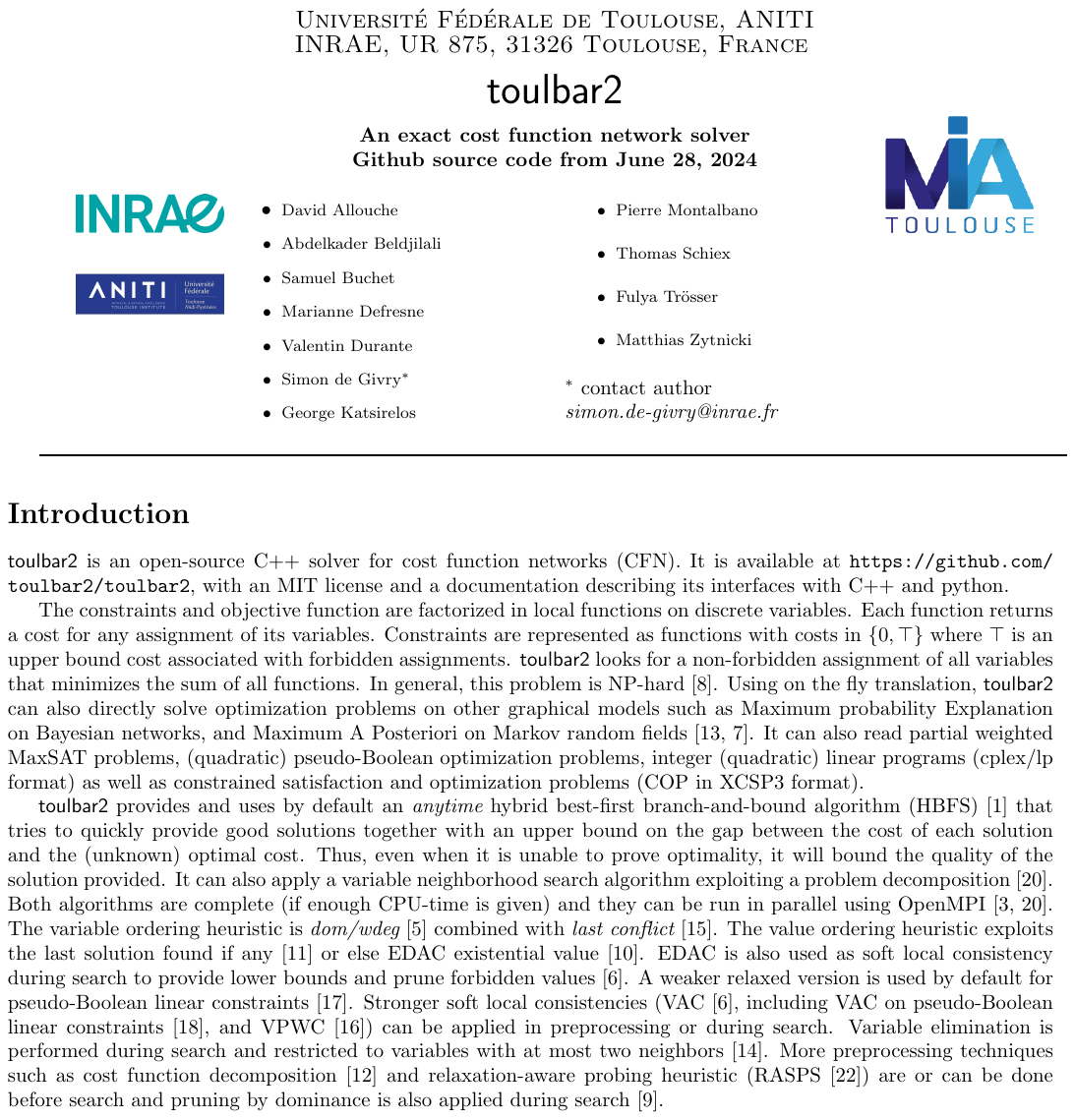}


\chapter{Results}

In this chapter, rankings for the various tracks of the \x3 Competition 2024 are given.
Importantly, remember that you can find all detailed results, including all traces of solvers at \href{https://www.cril.univ-artois.fr/XCSP24/}{https://www.cril.univ-artois.fr/XCSP24/}.

\section{Context}

\bigskip
Remember that the tracks of the competition are given by Table \ref{tab:anysolver} and Table \ref{tab:minisolver}.

\begin{table}[h!]
\begin{center}
\begin{tabular}{cccc} 
\toprule
\textcolor{dred}{\bf Problem} &  \textcolor{dred}{\bf Goal} &  \textcolor{dred}{\bf Exploration} &  \textcolor{dred}{\bf Timeout} \\
\midrule
CSP  & one solution & sequential & 40 minutes \\
COP  & best solution & sequential & 40 minutes \\
Fast COP  & best solution & sequential & 4 minutes \\
// COP  & best solution & parallel & 40 minutes \\
\bottomrule
\end{tabular}
\end{center}
\caption{Standard Tracks. \label{tab:anysolver}}
\end{table}

\begin{table}[h!]
\begin{center}
\begin{tabular}{cccc} 
\toprule
\textcolor{dred}{\bf Problem} &  \textcolor{dred}{\bf Goal} &  \textcolor{dred}{\bf Exploration} &  \textcolor{dred}{\bf Timeout} \\
\midrule
Mini CSP  & one solution & sequential & 40 minutes \\
Mini COP  & best solution & sequential & 40 minutes \\
\bottomrule
\end{tabular}
\end{center}
\caption{Mini-Solver Tracks. \label{tab:minisolver}}
\end{table}

\noindent Also, note that:

\begin{itemize}
\item The cluster was provided by CRIL and is composed of nodes with 2 quad-core Intel(R) Xeon(R) CPU E5-2643 0 @ 3.30GHz, each equipped with 32GiB RAM (24GiB for jobs). 
\item Each solver was allocated a CPU and 64 GiB of RAM, independently from the tracks.
\item Timeouts were set accordingly to the tracks through the tool \texttt{runsolver}:
 \begin{itemize}
    \item sequential solvers in the fast COP track were allocated 3 minutes of CPU time and 4.5 minutes of Wall Clock time,
    \item other sequential solvers were allocated 30 minutes of CPU time and 45 min of Wall Clock time,
    \item parallel solvers were allocated 4 CPU and 30 minutes of Wall Clock Time.
 \end{itemize}
 \item The selection of instances for the Standard tracks was composed of 200 CSP instances and 250~COP instances.
\item The selection of instances for the Mini-solver tracks was composed of 150 CSP instances and 150~COP instances.
\end{itemize}

\paragraph{About Scoring.}
The number of points won by a solver $S$ is decided as follows:
\begin{itemize}
\item for CSP, this is the number of times $S$ is able to solve an instance, i.e., to decide the satisfiability of an instance (either exhibiting a solution, or indicating that the instance is unsatisfiable)
\item for COP, this is, roughly speaking, the number of times $S$ gives the best known result, compared to its competitors.
More specifically, for each instance $I$:
\begin{itemize}
\item if $I$ is unsatisfiable, 1 point is won by $S$ if $S$ indicates that the instance $I$ is unsatisfiable, 0 otherwise,
\item if $S$ provides a solution whose bound is less good than another one (found by another competiting solver), 0 point is won by $S$,
\item if $S$ provides an optimal solution, while indicating that it is indeed the optimality, 1 point is won by $S$,
\item if $S$ provides (a solution with) the best found bound among all competitors, this being possibly shared by some other solver(s), while indicating no information about optimality: 1 point is won by $S$ if no other solver proved that this bound was optimal, $0.5$ otherwise.
\end{itemize}
\end{itemize}


\paragraph{Off-competition Solvers.} Some solvers were run while not officially entering the competition: we call them {\em off-competition} solvers.
\ace is one of them because its author (C. Lecoutre) conducted the selection of instances, which is a very strong bias.
Also, when two or more variants (by the same competiting team) of a same solver compete in a same track, only the best one is ranked (and the other ones considered as being off-competition).
To determine which solver variant is the best for a team, we compute the ranking with only the variants of the team competiting between them. 
This is why, for example, in some cases, solvers submitted by the CPMpy team were discarded from the ranking. 

\section{Rankings}

Recall that, concerning ranking, two new rules are used when necessary:
\begin{itemize}
\item In case a team submits the same solver to both the main track and the mini-track for the same problem (CSP or COP), 
the solver will be ranked in the mini-track only if the solver is not one of the three best solvers in the main track.
\item In case several teams submit variations of the same solver to the same track, only
the team who developed the solver and the best other team with that solver will
be ranked (possibly, a second best other team, if the jury thinks that it is relevant)
\end{itemize}

\bigskip
\noindent The algorithm used in practice for establishing the ranking is:
\begin{enumerate}
\item first, off-competition solvers are discarded (this is the case for ACE in 2024)
\item second, in mini-tracks, solvers that are ranked 1st, 2nd or 3rd in the corresponding main track are discarded 
\item  third, any less efficient variation of the same solver (submitted by the same team) is discarded ; the jury considered that technologies behind CPMpy\_ortools, CPMpy\_chuffed and CPMpy\_gurobi are different enough to not apply this rule 
\end{enumerate}

\bigskip
Here are the rankings\footnote{The images of medals come from \href{https://freesvg.org/gold-medal-juhele-final}{freesvg.org}} for the 6 tracks.
\bigskip

\begin{minipage}{0.4\textwidth}
\begin{center}
  \begin{tabular}{|lcp{2.7cm}|}
    \hline
     \vspace{-0.2cm} &  &  \\
    \multirow{5}{2.05cm}{{\bf {\large ~ CSP}}} & \includegraphics[scale=0.15]{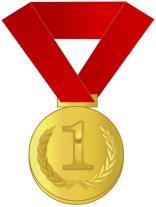} &  \vspace{-0.6cm} {\large Picat} \\
    &  & \\
    & \includegraphics[scale=0.15]{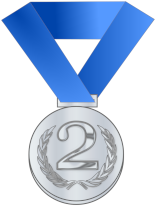} &  \vspace{-0.6cm} {\large CPMpy\_ortools} \\
    &  & \\
    & \includegraphics[scale=0.15]{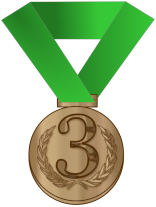} &  \vspace{-0.6cm} {\large Fun-sCOP} \\
    \hline 
  \end{tabular}
\end{center}
\end{minipage} \hspace{1.4cm}
\begin{minipage}{0.4\textwidth}
\begin{center}
  \begin{tabular}{|lcp{2.7cm}|}
    \hline
    \vspace{-0.2cm} & &  \\
    \multirow{5}{2.05cm}{{\bf {\large ~ COP}}} & \includegraphics[scale=0.15]{gold.png} &  \vspace{-0.6cm}  {\large CPMpy\_ortools} \\
    &  & \\
    & \includegraphics[scale=0.15]{silver.png} &  \vspace{-0.6cm} {\large Picat} \\
    &  & \\
    & \includegraphics[scale=0.15]{bronze.png} &  \vspace{-0.6cm} {\large CoSoCo} \\
    \hline
  \end{tabular}
\end{center}
\end{minipage}

\bigskip

\begin{minipage}{0.4\textwidth}  
\begin{center}
  \begin{tabular}{|lcp{2.7cm}|}
    \hline
    \vspace{-0.2cm} &  &  \\
    \multirow{5}{2.05cm}{{\bf {\large Fast COP}}} & \includegraphics[scale=0.15]{gold.png} &  \vspace{-0.6cm} {\large Picat} \\
    &  & \\
    & \includegraphics[scale=0.15]{silver.png} &  \vspace{-0.6cm} {\large CoSoCo\_fast} \\
    &  & \\
    & \includegraphics[scale=0.15]{bronze.png} &  \vspace{-0.6cm} {\large Choco} \\
    \hline
  \end{tabular}
\end{center}
\end{minipage} \hspace{1.4cm}
\begin{minipage}{0.4\textwidth}
\begin{center}
  \begin{tabular}{|lcp{2.7cm}|}
    \hline
    \vspace{-0.2cm} &  &  \\
    \multirow{5}{2.05cm}{{\bf {\large // COP}}} & \includegraphics[scale=0.15]{gold.png} &  \vspace{-0.6cm} {\large CPMpy\_ortools} \\
    &  & \\
    & \includegraphics[scale=0.15]{silver.png} &  \vspace{-0.6cm} {\large Choco} \\
    &  & \\
    & \includegraphics[scale=0.15]{bronze.png} &  \vspace{-0.6cm} {\large Toulbar2}  \\
    \hline
  \end{tabular}
\end{center}
\end{minipage}

\bigskip

\begin{minipage}{0.4\textwidth}  
\begin{center}
  \begin{tabular}{|lcp{2.7cm}|}
    \hline
    \vspace{-0.2cm} &  &  \\
    \multirow{5}{2.05cm}{{\bf {\large Mini CSP}}} &\includegraphics[scale=0.15]{gold.png} &  \vspace{-0.6cm} {\large CPMpy\_chuffed} \\
    &  & \\
    & \includegraphics[scale=0.15]{silver.png} &  \vspace{-0.6cm} {\large miniBTD} \\
    &  & \\
    & \includegraphics[scale=0.15]{bronze.png} &  \vspace{-0.6cm} {\large Nacre} \\
    \hline
  \end{tabular}
\end{center}
\end{minipage} \hspace{1.4cm}
\begin{minipage}{0.4\textwidth}
\begin{center}
  \begin{tabular}{|lcp{2.7cm}|}
    \hline
    \vspace{-0.2cm} &  &  \\
    \multirow{5}{2.14cm}{{\bf {\large Mini COP}}} & \includegraphics[scale=0.15]{gold.png} &  \vspace{-0.6cm} {\large Exchequer} \\
    &  & \\
    & \includegraphics[scale=0.15]{silver.png} &  \vspace{-0.6cm} {\large miniRBO} \\
    &  & \\
    & \includegraphics[scale=0.15]{bronze.png} &  \vspace{-0.6cm} {\large Toulbar2} \\
    &  &  \vspace{-0.6cm} {\large CPMpy\_gurobi} \\
    \hline
  \end{tabular}
\end{center}
\end{minipage}


\begin{thebibliography}{10}

\bibitem{compet22}
G.~Audemard, C.~Lecoutre, and E.~Lonca.
\newblock Proceedings of the 2022 {XCSP3} competition.
\newblock Technical Report arXiv:2209.00917, CoRR, 2022.
\newblock
  \href{https://arxiv.org/abs/2209.00917}{https://arxiv.org/abs/2209.00917}.

\bibitem{compet23}
G.~Audemard, C.~Lecoutre, and E.~Lonca.
\newblock Proceedings of the 2023 {XCSP3} competition.
\newblock Technical Report arXiv:2312.05877, CoRR, 2023.
\newblock
  \href{https://arxiv.org/abs/2312.05877}{https://arxiv.org/abs/2312.05877}.

\bibitem{BBSS_local}
G.~Belov, N.~Boland, M.~Savelsbergh, and P.~Stuckey.
\newblock Local search for a cargo assembly planning problem.
\newblock In {\em Proceedings of CPAIOR'14}, pages 159--175, 2014.

\bibitem{BCG_bwinning}
E.~Berlekamp, J.~Conway, and R.~Guy.
\newblock {\em Winning ways for your mathematical plays, vol. 2: games in
  particular}.
\newblock Academic Press, 1982.

\bibitem{xcsp3}
F.~Boussemart, C.~Lecoutre, G.~Audemard, and C.~Piette.
\newblock {XCSP3:} an integrated format for benchmarking combinatorial
  constrained problems.
\newblock Technical Report arXiv:1611.03398, CoRR, 2016.
\newblock
  \href{https://arxiv.org/abs/1611.03398}{https://arxiv.org/abs/1611.03398}.

\bibitem{xcsp3core}
F.~Boussemart, C.~Lecoutre, G.~Audemard, and C.~Piette.
\newblock {XCSP3}-core: {A} format for representing constraint
  satisfaction/optimization problems.
\newblock Technical Report arXiv:2009.00514, CoRR, 2020.
\newblock
  \href{https://arxiv.org/abs/2009.00514}{https://arxiv.org/abs/2009.00514}.

\bibitem{DLVK_fast}
A.~Deza, C.~Liu, P.~Vaezipoor, and E.~Khalil.
\newblock Fast matrix multiplication without tears: A constraint programming
  approach.
\newblock In {\em Proceedings of CP'23}, pages 14:1--14:15, 2023.

\bibitem{DDa_sabio}
R.~Duque, J.-F. Díaz, and A.~Arbelaez.
\newblock {SABIO: An Implementation of MIP and CP for Interactive Soccer
  Queries}.
\newblock In {\em Proceedings of CP'16}, pages 575--583, 2016.

\bibitem{ENT_solving}
K.~Easton, G.~Nemhauser, and M.~Trick.
\newblock Solving the travelling tournament problem: A combined integer
  programming and constraint programming approach.
\newblock In {\em Proceedings of PATAT'02}, pages 100--112, 2002.

\bibitem{EMZ_scalable}
M.~Erascu, F.~Micota, and D.~Zaharie.
\newblock Scalable optimal deployment in the cloud of component-based
  applications using optimization modulo theory, mathematical programming and
  symmetry breaking.
\newblock {\em Journal of Logical and Algebraic Methods in Programming},
  121:100664, 2021.

\bibitem{GJMN_generating}
I.~Gent, C.~Jefferson, I.~Miguel, and P.~Nightingale.
\newblock Generating special-purpose stateless propagators for arbitrary
  constraints.
\newblock In {\em Proceedings of CP'10}, pages 206--220, 2010.

\bibitem{JMMT_modelling}
C.~Jefferson, A.~Miguel, I.~Miguel, and A.~Tarim.
\newblock Modelling and solving english peg solitaire.
\newblock {\em Computers \& Operations Research}, 33(10):2935--2959, 2006.

\bibitem{LV_branch}
E.~Lam and P.~Van Hentenryck.
\newblock A branch-and-price-and-check model for the vehicle routing problem
  with location congestion.
\newblock {\em Constraints}, 21(3):394--412, 2016.

\bibitem{ace}
C.~Lecoutre.
\newblock {ACE}, a generic constraint solver.
\newblock Technical Report arXiv:2302.05405, CoRR, 2023.
\newblock
  \href{https://arxiv.org/abs/2302.05405}{https://arxiv.org/abs/2302.05405}.

\bibitem{pycsp3}
C.~Lecoutre and N.~Szczepanski.
\newblock {PyCSP$^3$}: Modeling combinatorial constrained problems in {Python}.
\newblock Technical Report arXiv:2009.00326, CoRR, 2020.
\newblock
  \href{https://arxiv.org/abs/2009.00326}{https://arxiv.org/abs/2009.00326}.

\bibitem{MS_book}
K.~Marriott and P.~Stuckey.
\newblock {\em Programming with Constraints}.
\newblock MIT Press, 1998.

\bibitem{MSS_solver}
N.~Musliu, A.~Schutt, and P.~Stuckey.
\newblock Solver independent rotating workforce scheduling.
\newblock In {\em Proceedings of CPAIOR'18}, pages 429--445, 2018.

\bibitem{RCP23}
S.~Roussel, T.~Polacsek, and A.~Chan.
\newblock Assembly line preliminary design optimization for an aircraft.
\newblock In {\em Proceedings of CP'23}, pages 32:1--32--19, 2023.

\bibitem{SS_cargo}
M.~Savelsbergh and O.~Smith.
\newblock Cargo assembly planning.
\newblock {\em EURO Journal on Transportation and Logistics}, 4(3):321--354,
  2015.

\bibitem{XBHL_random}
K.~Xu, F.~Boussemart, F.~Hemery, and C.~Lecoutre.
\newblock Random constraint satisfaction: easy generation of hard (satisfiable)
  instances.
\newblock {\em Artificial Intelligence}, 171(8-9):514--534, 2007.

\bibitem{DBLP:conf/padl/ZhouK16}
Neng{-}Fa Zhou and H{\aa}kan Kjellerstrand.
\newblock The {Picat-SAT} compiler.
\newblock In {\em Proceedings of PADL'16}, pages 48--62. Springer, 2016.

\end{thebibliography}

\end{document}